\documentclass[conference]{IEEEtran}
\IEEEoverridecommandlockouts
\usepackage{cite}
\usepackage{amsmath,amssymb,amsfonts}
\usepackage{algorithmic}
\usepackage{graphicx}
\usepackage{textcomp}
\usepackage[keeplastbox]{flushend}
\usepackage{mathtools}
\usepackage{multirow}
\usepackage[linesnumbered,ruled,vlined]{algorithm2e}
\usepackage{tabularx}
\usepackage{hhline}

\usepackage{rotating}
\usepackage{tabularx}

\begin{document}


\title{A Survey of Complex-Valued Neural Networks}


\author{\IEEEauthorblockN{Joshua~Bassey,~Xiangfang Li,~Lijun Qian}
\IEEEauthorblockA{Center of Excellence in Research and Education for Big Military Data Intelligence (CREDIT) \\ 
Department of Electrical and Computer Engineering \\
Prairie View A\&M University, Texas A\&M University System  \\
Prairie View, TX 77446, USA \\
Email: jbassey@student.pvamu.edu, xili@pvamu.edu, liqian@pvamu.edu}
}


\maketitle

\begin{abstract}
Artificial neural networks (ANNs) based machine learning models and especially deep learning models have been widely applied in computer vision, signal processing, wireless communications, and many other domains, where complex numbers occur either naturally or by design. However, most of the current implementations of ANNs and machine learning frameworks are using real numbers rather than complex numbers. There are growing interests in building ANNs using complex numbers, and exploring the potential advantages of the so called complex-valued neural networks (CVNNs) over their real-valued counterparts. In this paper, we discuss the recent development of CVNNs by performing a survey of the works on CVNNs in the literature. Specifically, detailed review of various CVNNs in terms of activation function, learning and optimization, input and output representations, and their applications in tasks such as signal processing and computer vision are provided, followed by a discussion on some pertinent challenges and future research directions.

\end{abstract}

\begin{IEEEkeywords}
complex-valued neural networks; complex number; machine learning; deep learning
\end{IEEEkeywords}

\section{Introduction}
Artificial neural networks (ANNs) are data-driven computing systems inspired by the dynamics and functionality of the human brain. With the advances in machine learning especially in deep learning, ANNs based  deep learning models  have gain tremendous usages in many domains and have been tightly fused into our daily lives. Applications such as automatic speech recognition make it possible to have conversations with computers, enable computers to generate speech and musical notes with realistic sounds, and separate a mixture of speech into single audio-streams for each speaker~\cite{Saroff2018}. Other examples include object identification and tracking, personalized recommendations, and automating important tasks more efficiently~\cite{Shinde2018}.
 
In many of the practical applications, complex numbers are often used such as in telecommunications, robotics, bioinformatics, image processing, sonar, radar, and speech recognition. This suggests that ANNs using complex numbers to represent inputs, outputs, and parameters such as weights, have potential in these domains. For example, it has been shown that the phase spectra is able to encode fine-scale temporal dependencies~\cite{Saroff2018}. Furthermore, the real and imaginary parts of a complex number have some statistical correlation. By knowing in advance the importance of phase and magnitude to our learning objective, it makes more sense to adopt a complex-valued model, as this offers a more constrained system than a real-valued model~\cite{Hirose2011}.

Complex-valued neural networks (CVNN) are ANNs that process information using complex-valued parameters and variables~\cite{HIROSE1994}.  The main reason for their advocacy lies in the difference between the representation of the arithmetic of complex numbers, especially the multiplication operation. In other words, multiplication function which results in a phase rotation and amplitude modulation yields an advantageous reduction of the degree of freedom~\cite{Hirose1992}. The advantage of ANNs is their self-organization and high degree of freedom in learning. By knowing \textit{a priori} about the amplitude and phase portion in data, a potentially dangerous portion of the freedom can be minimized by using CVNNs. 

Recently, CVNNs have received increased interests in signal processing and machine learning research communities. In this paper, we discuss the recent development of CVNNs by performing a survey of the works on CVNNs in the literature.  The contributions of this paper include
\begin{enumerate}
\item A systematic review and categorization of the state-of-the-art CVNNs has been carried out based on their activation functions, learning and optimization methods, input and output representations, and their applications in various tasks such as signal processing and computer vision. 
\item Detailed description of the different schools of thoughts, similarities and differences in approaches, and advantages and limitations of various CVNNs are provided. 
\item Further discussions on some pertinent challenges and future research directions are given.
\end{enumerate}
To the best of our knowledge, this is the first work solely dedicated to a comprehensive review of complex-valued neural networks.

The rest of this paper is structured as follows. A background on CVNNs, as well as their use cases are presented in Section~\ref{sec:background}. Section~\ref{sec:AF} discusses CVNNs according to the type of activation functions used. Section~\ref{sec:learning} reviews  CVNNs based on their learning and optimization approaches. The CVNNs characterized by their input and output representations are reviewed in Section~\ref{sec:in_out}. Various applications of CVNNs are presented in Section~\ref{sec:applications} and challenges and potential research directions are discussed in Section~\ref{sec:challenges}. Section~\ref{sec:conclusion} contains the concluding remarks.

The symbols and notations used in this review are summarized in Table~\ref{table:notations}.

\begin{table}
\caption{Symbols and Notations}
\begin{tabular}{ll}
C & multivalued neural network (MVN) learning rate \tabularnewline
$\mathbb{C}$ & complex domain \tabularnewline
$\mathbb{R}$ & real domain \tabularnewline
$d$ & desired output \tabularnewline
$e$ & individual error of network output \tabularnewline
$e_{log}$ & logarithmic error \tabularnewline
$E$ & error of network output \tabularnewline
$f$ & activation function \tabularnewline
$i$ & imaginary unity \tabularnewline
$Im$ & imaginary component \tabularnewline
$j$ & values of k-valued logic \tabularnewline
$J$ & regularization cost function \tabularnewline
$k$ & output indices  \tabularnewline
$l$ & indices of preceeding network layer  \tabularnewline
$n$ & indices for input samples \tabularnewline
$N$ & total number of input samples \tabularnewline
$o$ & actual output (prediction) \tabularnewline
p & dimension of real values \tabularnewline
$Re$ & real component \tabularnewline
$m$ & indices for output layer of MVN \tabularnewline
t & regularization threshold parameter  \tabularnewline
T & target for MVN \tabularnewline 
$u$ & real part of activation function  \tabularnewline
$v$ & imaginary part of activation function  \tabularnewline
$x$ & real part of weighted sum  \tabularnewline
$y$ & imaginary part of weighed sum  \tabularnewline
$Y$ & output of MVN \tabularnewline
$\delta$ & partial differential  \tabularnewline
$\Delta$ & total differential  \tabularnewline
$\nabla$ & gradientoperator  \tabularnewline
$\mathfrak{l}(e)$ & mean square loss function\tabularnewline
$\mathfrak{l}(e_{log})$ & logarithmic loss function \tabularnewline
$\epsilon ^{*}$ & global error for MVN  \tabularnewline
$\epsilon $ & neuron error for MVN  \tabularnewline
$\omega$ & error threshold for MVN \tabularnewline
$\hat{\beta}$ & regularized weights  \tabularnewline
$\lambda$ & regularization parameter \tabularnewline
$\mathbf{X}$ & all inputs \tabularnewline
$w$ & individual weight \tabularnewline
$\mathbf{W}$ & all network weights \tabularnewline
$z$ & weighted sum \tabularnewline
$\vert \cdot \vert$ & modulo operation \tabularnewline
$\lVert \cdot \rVert$ & euclidean distance \tabularnewline
$\angle$ & angle \tabularnewline
\end{tabular}
\label{table:notations}
\end{table}

\section{Background}
\label{sec:background}
\subsection{Historical Notes}
\label{subsec:history}
The ADALINE machine~\cite{Widrow1960}, a one-neuron, one-layer machine is one of the earliest implementations of a trainable neural network influenced by the Rosenblatt perceptron~\cite{Rosenblatt1958}. ADALINE used least mean square (LMS) and stochastic gradient descent for deriving optimal weights. 

The LMS was first extended to the complex domain in~\cite{Widrow1975}, where gradient descent was derived with respect to the real and imaginary part. Gradient descent was further generalized in~\cite{Brandwood1983} using Wirtinger Calculus such that the gradient was performed with respect to complex variables  instead of the real components. Wirtinger Calculus provides a framework for obtaining the gradient with respect to complex-valued functions~\cite{Wirtinger1927}. The complex-valued representation of the gradient is equivalent  to obtaining the gradients of the real and imaginary components in part.

\subsection{Why Complex-Valued Neural Networks}
\label{subsec:CVNN_case}
Artificial neural networks (ANNs) based machine learning models and especially deep learning models have gained wide spread usage in recent years. However, most of the current implementations of ANNs and machine learning frameworks are using real numbers rather than complex numbers. There are growing interests in building ANNs using complex numbers, and exploring the potential advantages of the so called complex-valued neural networks (CVNNs) over their real-valued counterparts. The first question is: why CVNNs are needed?

Although in many analyses involving complex numbers, the individual components of the complex number have been treated independently as real numbers, it would be erroneous to apply the same concept to CVNNs by assuming that a CVNN is equivalent to a two-dimensional real-valued neural network. In fact, it has been shown that this is not the case~\cite{Hirose2012}, because the operation of complex multiplication limits the degree of freedom of the CVNNs at the synaptic weighting. This suggests that the phase-rotational dynamics strongly underpins the process of learning.

From a biological perspective, the complex-valued representation has been used in a neural network~\cite{Reichert2014}. The output of a neuron was expressed as a function of its firing rate specified by its amplitude, and the comparative timing of its activity is represented by its phase. Exploiting complex-valued neurons resulted in more versatile representations. With this formulation, input neurons with similar phases add constructively and are termed \textit{synchronous}, and \textit{asynchronous} neurons with dissimilar phases interfere with each other because they add destructively. This is akin to the behavior of the gating operation applied in deep feedforward neural networks~\cite{Srivastava2015}, as well as in recurrent neural networks~\cite{Cho2014}. In the gating mechanism, the controlling gates are typically the sigmoid-based activation, and synchronization describes the propagation of inputs with simultaneously high values held by their controlling gates. This property of incorporating phase information may be one of the reasons for the effectiveness of using complex-valued representations in recurrent neural networks.

The importance of phase is backed from the biological perspective and also from a signal processing point of view. Several studies have shown that the intelligibility of speech is affected largely by the information contained in the phase portion of the audio signal~\cite{Shi2006,Saroff2018}. Similar results have also been shown for images. For example,  it was shown in~\cite{Oppenheim1981} that by exploiting the information encoded in the phase of an image, one can sufficiently recover most of the information encoded in its magnitude. This is because the phase describes objects in an image in terms of edges, shapes and their orientations.

From a computational perspective, holographic reduced representations (HRRs) were combined to enhance the storage of data as key-value pairs in~\cite{Danihelka2016}.  The idea is to mitigate two major limitations of recurrent neural networks: (1) the dependence of the number of memory cells on the recurrent weight matrices' size, and (2) lack of memory indexing during writing and reading, causing the inability to learn to represent common data structures such as arrays. The complex conjugate was used for key retrieval instead of the inverse of the weight matrix in~\cite{Danihelka2016}. 
The authors showed that the use of complex numbers by Holographic Reduced Representation for data retrieval from associative memories is more numerically stable and efficient. They further showed that even conventional networks such as residual networks~\cite{He2016} and Highway networks~\cite{Srivastava2015} display similar framework to that of associative memories. In other words, each residual network uses the identity connection to insert or ``aggregate'' the computed residual into memory.

Furthermore, orthogonal weight matrices have been shown to mitigate the well-known problem of exploding and vanishing gradient problems associated with recurrent neural networks in the real-valued case. Unitary weight matrices are a generalization of orthogonal weight matrices to the complex plane. Unitary matrices are the core of Unitary RNNs~\cite{Arjovsky2016}, and uncover spectral representations by applying the discrete Fourier transform. Hence, they offer richer representations than orthogonal matrices. This idea behind unitary RNNs was exploited in~\cite{Wisdom2016}, where a general framework was derived for learning unitary matrices and applied on a real-world speech problem as well as other toy tasks.

As for the theoretical point of view, a complex number can be represented either in vector or matrix form. Consequently,  the multiplication of two complex numbers can be represented as a matrix-vector multiplication. However, the use of the matrix representation increases the number of dimensions and parameters of the model. It is also common knowledge in machine learning that the more complicated a model in terms of parameters, the greater the tendency of the model to overfit. Hence, using real-valued operations to approximate these complex parameters could result in a model with undesirable generalization characteristics. On the contrary, in the complex domain, the matrix representation mimics a rotation matrix. This means that half of the entries of the matrix is fixed once the other half is known. This constraint reduces the degrees of freedom and enhances the generalization capacity of the model. 

Based on the above discussions, it is clear that there are two main reasons for the use of complex numbers for neural networks 
\begin{enumerate}
\item In many application domains such as wireless communication or audio processing, where complex numbers occur naturally or by design, there is a correlation between the real and imaginary parts of the complex signal. For instance, Fourier transform involves a linear transformation, with a direct correspondence between the multiplication of a signal by a scalar in the time-domain, and multiplying the magnitude of the signal in the frequency domain. In the time domain, the circular rotation of a signal is equivalent to shifting its phase in the frequency domain. This means that during phase change, the real and imaginary components of a complex number are statistically correlated. This assumption is voided when real-valued models are used especially on the frequency domain signal. 

\item If the relevance of the magnitude and phase to the learning objective is known \emph{a priori}, then it is more reasonable to use a complex-valued model because it  imposes more constrains on the complex-valued model than a real-valued model would. \end{enumerate}

\section{Activation Functions of CVNNs}
\label{sec:AF}
Activation functions introduce non-linearity to the affine transformations in neural networks. This gives the model more expressiveness. Given an input $x \in \mathbb{C}^{M}$, and weights $W \in \mathbb{C}^{N \times M}$, where $M$ and $N$ represent the dimensionality of the input and output respectively, the output $y \in \mathbb{C}^{N}$ of any neuron is:

\begin{eqnarray}
f(z) &=& \mathbf{Wx},
\end{eqnarray}
where $f$ is a nonlinear activation-function operation performed element-wise. Neural networks have been shown to be neural approximators using the sigmoid squashing activation functions that are monotonic as well as bounded~\cite{Cybenko1989,Barron1994,FUNAHASHI1989, HORNIK1989}.

\begin{table}[!t]
\caption{Complex-Valued Activation Functions}
\label{table:activation_functions}
\begin{tabular}{|p{3.4cm}|p{4.6cm}|}
 \hline
{\bf Activation Function} & {\bf Corresponding Publications} \\
\hhline{|=|=|}
Split-type A  & \cite{Benvenuto1992,Hayakawa2018,ACAR2018,Ishizuka2018,Yi2018a,Cevik2018,Gleich2018,Gong2017,Zhang2017b,Popa2017a,Liu2017,
Peker2016,Amilia2015,Maeda2014,Liu2014,Hirose2012,Olanrewaju2011,Haensch2010,Nait2010,Kitajima2010,Ogawa2008,Amin2008,
Nishikawa2005,Yadav2005,Kataoka1998,Kinouchi1996,Tan2020}\\\hline

Split-type B  & \cite{Georgiou1992,Hirose1992,Hayakawa2018,Hu2018a,Hu2018b,Hata2016,Ding2014,Suzuki2013,
Mizote2013,Al_Nuaimi2012,Hirose2012,Hu2008,Nishikawa2006,Nishikawa2005}  \\\hline

Fully Complex (ETF) & \cite{Nitta2018a,Kominami2017,Mandic2009,Hanna2002,Adali2000,Georgiou1992,Kim2002}  \\\hline

Non Parametric & \cite{Scardapane2018}  \\\hline

Energy Functions & \cite{Kobayashi2019,Popa2018a,Kobayashi2018c,Kuroe2002}  \\\hline

Complex ReLU & \cite{Yuan2019,Li2019,Gao2019,Gu2018,Matlacz2018,Popa2018b,Shafran2018,Popa2017a,Lee2017,Lee2017b}  \\\hline

Nonlinear Phase & \cite{Kobayashi2019b,Aizenberg2018a,Aizenberg2018b,Kobayashi2018b,Virtue2017,Ronghua2017,Aizenberg2017a,Kobayashi2017,
Aizenberg2016a,Hacker2016,Catelani2016,Aizenberg2014a,Pavaloiu2014,Valle2014,Aizenberg2013a,Wu2013,
Chen2013,Aizenberg2011a,Aizenberg2008b,Donq2001,Aizenberg2000,Aizenberg2000b,Aizenberg1996,Olga2014,Jankowski1996,Tanaka2009,
Aizenberg2006}  \\\hline

Linear Activation  & \cite{Ding2019}  \\\hline

Split Kernel Activation  & \cite{Scardapane2018}  \\\hline

Complex Kernel Activation  & \cite{Scardapane2018}  \\\hline

Absolute Value Activation & \cite{Marseet2017}  \\\hline

Hybrid Activation  & \cite{Wilmanski2016}  \\\hline

Mobius Activation  & \cite{Wilmanski2016}  \\
\hhline{|=|=|}
\end{tabular}

\end{table}

The majority of the activation functions that have been proposed for CVNNs in the literature are summarized in Table \ref{table:activation_functions}. Activation functions are generally thought of as being holomorphic or non-holomorphic. In other words, the major concern is whether the activation function is differentiable everywhere, differentiable around certain points, or not differentiable at all. Complex functions that are holomorphic at every point are known as ``\textit{entire functions}''. However, in the complex domain, one cannot have a bounded complex activation function that is complex-differentiable at the same time. This stems from Liouville's theorem which states that all bounded entire functions are a constant. Hence, it is not possible to have a CVNN that uses squashing activation functions and is entire.

\begin{figure}
	 \centering
    	 \includegraphics[width=0.4\textwidth]{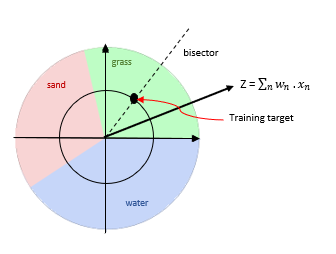}
      	\caption{Geometric Interpretation for discrete-valued MVN activation function}
    \label{fig:MVNdiscAF}
\end{figure}

One of the first works on complex activation function was done by Naum Aizenberg~\cite{Aizenberg1973, Ajzenberg1972}. According to these works, \textit{A multi-valued neuron (MVN) is a neural element with n inputs and one output lying on the unit circle, and with complex-valued weights}. The mapping is described as:

\begin{equation}
f(x_{1},...,x_{n}) = f(w_{0} + w_{1}x_{1} +...+ w_{n}x_{n})
\label{eq:cvmap}
\end{equation}
where $x_{1},...,x_{n}$ are the dependent variables of the function, and $w_{0}, w_{1},..,w_{n}$ are the weights. All variables are complex, and all outputs of the function are the \textit{kth} roots of unity $\epsilon^{j}$ = exp$(i2\pi j/k), j \in [0, k-1]$, and $i$ is imaginary unity.
$f$ is an activation function on the weighted sum defined as:
\begin{eqnarray}\nonumber
f(z) &=& exp(i2\pi j/k),\\
&   \textit{if} & 2\pi j/k \leq arg(z) < 2\pi(j+1)/k 
\label{eq:discAF}
\end{eqnarray}
where $w_{0} + w_{1}x_{1} +...+ w_{n}x_{n}$ is the weighted sum, $arg(z)$ is the argument of $z$, and $j=0,1,...,k-1$ represents values of the k-valued logic. A geometric interpretation of the MVN activation function is shown in Figure \ref{fig:MVNdiscAF}. The activation function represented by equation (\ref{eq:discAF}) divides the complex plain into \textit{k} equal sectors, and  implements a mapping of the entire complex plane onto the unit circle. 

The same concept can be extended to continuous-valued inputs. By making $k \rightarrow \infty$, the angles  of the sectors in Figure \ref{fig:MVNdiscAF} will tend towards zero. This continuous-valued MVN is obtained by transforming equation (\ref{eq:discAF}) to:

\begin{equation}
f(z) = exp(i(arg(z))) = e^{iArg(z)} = z / \vert z \vert
\label{eq:contAF}
\end{equation}
where z is the weighted sum, $\vert z \vert$ is the modulo of $z$. Equation (\ref{eq:discAF}) maps the complex plane to a discrete subset of points on the unit circle whereas equation (\ref{eq:contAF}) maps the complex plane to the entire unit circle. 

\begin{figure}
{\includegraphics[width = 3.4in]{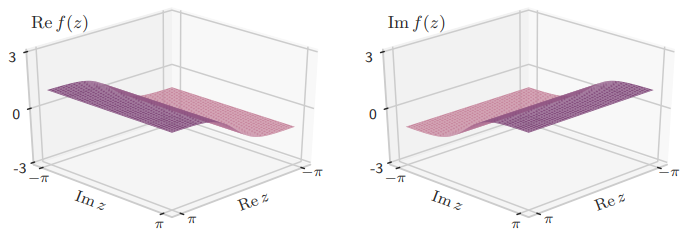}} 
\caption{Split-type hyperbolic tangent activation function}
\label{fig:split_tanhAF}
\end{figure}

There is no general consensus on the most appropriate activation function for CVNNs in the literature. The main requirement is to have a nonlinear function that is not susceptible to exploding or vanishing gradients during training. Since the work of Aizenberg, other activation functions have been proposed. For example, Holomorphic functions~\cite{Clarke1990} have been proposed and used in so called ``fully complex'' networks. The hyperbolic tangent is an example of a fully complex activation function and has been used in~\cite{Mandic2009}. Figure \ref{fig:split_tanhAF} shows its surface plots. It can be observed that singularities in the output can be caused by values on the imaginary axis. In order to avoid explosion of values, inputs have to be properly scaled.

Some researchers suggest that it is not necessary to impose the strict constraint of requiring that the activation function be holomorphic. Those of this school of thought advocate for activations which are differentiable with respect to their imaginary and real parts. These activations are called \textit{split activation functions} and can either be real-imaginary (Type A) or amplitude-phase (Type B)~\cite{Kuroe2003}. A Type A split-real activation was used in~\cite{Benvenuto1992} where the real and imaginary parts of the signal were input into a Sigmoid function. A Type B split amplitude-phase nonlinear activation function that squashes the magnitude and preserves phase was proposed in~\cite{Georgiou1992} and~\cite{Hirose1992}. 

While Type B activations preserve the phase and squashes the magnitude, activation functions that instead preserve the magnitude and squash the phase have been proposed. These type of activation functions termed ``phasor networks'' when originally introduced~\cite{Noest1988}, where output values extend from the origin to the unit circle~\cite{Jankowski1996,Miyajima2000,Hirose1992b,Nemoto2002,Zemel1995}. The multi-threshold logic activation functions used by multi-valued neural networks~\cite{Aizenberg1995} are also based on a similar idea.

Although most of the early approaches favor the split method based on non-holomorphic functions, gradient-preserving backpropagation can still be performed on fully complex activations. Consequently, elementary transcendental functions (ETF) have been proposed~\cite{Kim2001,Kim2002,Kim2003} to train the neural network when there are no singularities found in the domain of interest. This is because singularities in ETFs are isolated. 

Radial basis functions have also been proposed for complex networks~\cite{Inhyok1995, Chen1994a, Chen1994b,Deng2002}, and spline-based activation functions were proposed in works such as~\cite{Uncini1999} and~\cite{Scarpiniti2008}. In a more recent work~\cite{Scardapane2018}, non-parametric functions were proposed. The non-parametric kernels can be used in the complex domain directly or separately on the imaginary and real components.

Given its widespread adoption and success in deep real-valued networks, the ReLU activation~\cite{Hahnloser2000}:
\begin{equation}
ReLU(x) := max(x,0) \; ,
\end{equation}
has been proposed because it mitigates the problem of vanishing gradients encountered with Sigmoid activation. For example, ReLU was applied in~\cite{Arjovsky2016} after a trainable bias parameter was added to the magnitude of the complex number. In their approach, the phase is preserved while the magnitude is non-linearly transformed. The authors in~\cite{Trabelsi2017} applied ReLU separately on both real and imaginary parts of the complex number. The phase is nonlinearly mapped when the input lies in  quadrant other than the upper right quadrant of the Argand diagram. However, neither of these activations are holomorphic.

 The activation functions discussed in this section along with some others are listed in Table \ref{table:activation_functions}. For example, a cardioid activation in~\cite{Virtue2017}  is defined as:
\begin{equation}
f(z) = \frac{1}{2} (1 + cos(\angle z))z , 
\end{equation}
and this was applied in~\cite{Virtue2017} for magnitude resonance imaging (MRI) fingerprinting. The cardioid activation function is a phase sensitive complex extension of the ReLU. 

There are other activation functions besides those described in this section. 
There are activation functions which are suitable for both real and complex hidden neurons. There is still no agreed consensus on which scenarios warrant the use of holomorphic functions such as ETFs, or the use of nonholomorphic functions that are more closely related to the nonlinear activation functions widely used in current state-of-the-art real-valued deep architectures.
In general, there are no group of activation functions that are  deemed the best for both real or complex neural networks.

\section{Optimization and learning in CVNNs}
\label{sec:learning}
Learning in neural networks refers to the process of tuning the weights of the network to optimize learning objectives such  as minimizing a loss function. The optimal set of weights are those that allow the neural network generalize best to out-of-sample data. Given the desired and predicted output of a complex neural network, or ground truth and prediction in the context of supervised learning, $d \in \mathbb{C}^{N}$ and $o \in \mathbb{C}^{N}$, respectively, the error is
\begin{equation}
\label{eq:comp error}
e := d-o .
\end{equation}  

The complex mean square loss is a non-negative scalar
\begin{eqnarray}
\label{eq:comp_loss}
\mathcal{L}(e)  &=& \sum_{k=0}^{N-1} \vert e_{k} \vert ^{2} \\
&=& \sum_{k=0}^{N-1} e_{k} \bar{e}_{k} .
\end{eqnarray}
It is also a real-valued mapping that tends to zero as the modulus of the complex error reduces. 

The log error between the target and the prediction is given by~\cite{Savitha2013}
\begin{eqnarray}
\label{eq:log error}
(e_{log}) :=  \sum_{k=0}^{N-1} \left( \text{log } (o_{k}) - \text{log} (d_{k}) \right) \overline{\left( \text{log } (o_{k}) - \text{log} (d_{k}) \right)} 
\end{eqnarray}
where $\overline{(\cdot)}$ is the conjugate function. If $d = r \: \text{exp} (i \phi)$ and  $o = \hat{r} \: \text{exp} (i \hat{ \phi})$ are the polar representations of the desired and actual outputs respectively, the log error function  is a monotonically decreasing function. With this representation, the errors in magnitude and phase can have explicit representation in the logarithmic loss function given as:
\begin{equation}
\label{eq:comp_log_loss}
\mathcal{L}(e_{log}) =  \frac{1}{2} \bigg ( log \bigg [ \frac{\hat{r}_{k}}{r_{k}} \bigg ]^{2} + \big [ \hat{\phi}_{k} - \phi_{k} \big ]^{2} \bigg )
\end{equation}
such that $e_{log} \rightarrow 0$ when $\hat{r}_{k} \rightarrow r_{k}$ and $\hat{\phi}_{k} \rightarrow \phi_{k}$. The error functions in equations (\ref{eq:comp error}) and (\ref{eq:log error}) are suitable for complex-valued regression. For classification, one may map the output of the network to the real domain using a transform that is not necessarily holomorphic. 

Table \ref{table:learning_methods} details the most popular learning methods implemented in the literature. In general, there are two approaches for training CVNNs. The first approach follows the same method used in training the real-valued neural networks where the error is backpropagated from the output layer to the input layer using gradient descent. In the second approach, the error is backpropagated but \emph{without} gradient descent.

\begin{table}[!th]
\caption{Learning Methods for Complex-Valued Neural Networks}
\begin{tabular}{|p{4cm}|p{4cm}|}
\hline
{\bf Error Propagation Method} & {\bf Corresponding Publications} \\ 
\hhline{|=|=|}
Split-Real Backpropagation & \cite{Benvenuto1992,Widrow1960, Hirose1992,Cevik2018,Peker2016,Amilia2015,Liu2014,Hirose2012,Kitajima2010,Amin2008,Nishikawa2005,Yadav2005,Ding2014,Mizote2013
,Hanna2002,Yuan2019,Lee2017,Kim1990,Chen2000,Ceylan2005,Tay2007,Tsuzuki2013}\\\hline

Fully Complex Backpropagation (CR) & \cite{Widrow1975,Mandic2009,Georgiou1992,Kim2001,Adali2000,Kim2002,Zhang2017b,Popa2017a,Liu2017,Nitta2018a,Zhang2016,Popovic2004}  \\\hline

MLMVN  & \cite{Aizenberg1973,Ajzenberg1972,Aizenberg2007,Aizenberg1996,Aizenberg2018a,Ronghua2017,Aizenberg2016a,Hacker2016,Catelani2016,Aizenberg2014a,Pavaloiu2014,Aizenberg2013a,Chen2013,Aizenberg2011a,Aizenberg2008b,Aizenberg2000,Olga2014,Shamima2013,Aizenberg2008,Aizenberg2000c,Aizenberg2006}  \\\hline

Orthogonal Least Square & \cite{Chen1994a,Chen1994b} \\\hline

Quarternion-based Backpropagation & \cite{Hata2016} \\\hline

Hebbian Learning & \cite{Suzuki2013} \\\hline

Complex Barzilai-Borwein Training Method & \cite{Zhang2016} \\
\hhline{|=|=|}
\end{tabular}
\label{table:learning_methods}
\end{table}

\subsection{Gradient-based Approach}
Different approaches to the backpropagation algorithm in the complex domain were independently proposed by various researchers in the early 1990's. For example, a derivation for single hidden layer complex networks was given in~\cite{Kim1990}. In this work, the authors showed that a complex neural network with one hidden layer and Sigmoid activation was able to solve the XOR problem. Similarly using Sigmoid activation, the authors in~\cite{Leung1991} derived the complex backpropagation algorithm. Derivations were also given for Cartesian split activation function in~\cite{Benvenuto1992} and for non-holomorphic activation functions in~\cite{Georgiou1992}.

In reference to gradient based methods, the Wirtinger Calculus~\cite{Wirtinger1927} was used  to derive the complex gradient, Jacobian, and Hessian in~\cite{Brandwood1983} and~\cite{van1994}. Wirtinger  developed a framework  that simplifies the process obtaining the derivative of complex-valued functions with respect to both holomorphic and non-holomorphic functions. By doing this, the derivatives to the complex-valued functions can be computed completely in the complex domain instead of being computed with respect to the real and imaginary components independently. The Wirtinger approach has not always been favored, but it is beginning to gain more interest with newer approaches like~\cite{Al_Nuaimi2012,Haykin2014}.

The process of learning with complex domain backpropagation is similar to the learning process in the real domain. The error calculated after the forward pass is backpropagated to each neuron in the network, and the weights are adjusted in the backward pass. If the activation function of a neuron is $f(z) = u(x,y) + iv(x,y)$, where $z = x + iy$, $u$ and $v$ are the real and imaginary parts of $f$, and $x$ and $y$ are the real and imaginary parts of $z$. The partial derivatives $u_{x} = \partial u / \partial x, \; u_{y} = \partial u / \partial y, \; v_{x} = \partial v / \partial x, \; v_{y} = \partial v / \partial y$ are initially assumed to exist for all $z \in C$, so that the Cauchy-Riemann equations are satisfied. Given an input pattern, the error is given by
\begin{equation}
E = \frac{1}{2} \sum_{k} e_{k} \bar{e}_{k}, \qquad e_{k} = d_{k} - o_{k}
\end{equation}
where $d_{k}$ and $o_{k}$ are the desired and actual outputs of the $k$th neuron, respectively. The over-bar denotes the complex conjugate operation. Given a neuron $j$ in the network, the output $o_{j}$ is given by
\begin{equation}
o_{j} = f(z_{j}) = u^{j} + iv^{j}, \quad z_{j} = x_{j} + iy_{j} = \sum_{i=1} W_{jl}X_{jl}
\end{equation}
where the $W_{jl}$'s are the complex weights of neuron $j$ and $X_{jl}$ its complex input. A complex bias (1,0) may be added. The following partial derivatives are defined:
\begin{eqnarray}\nonumber
\frac{\partial x_{j}}{\partial W_{jlR}} = X_{jlR}, \; \frac{\partial y_{j}}{\partial W_{jlR}}= X_{jlI}, \;\\
 \frac{\partial x_{j}}{\partial W_{jlI}} = -X_{jlI}, \; \frac{\partial y_{j}}{\partial W_{jlI}} = X_{jlR}
\label{eq:partial_derivatives}
\end{eqnarray}
where $R$ and $I$ represent the real and imaginary parts. The chain rule is used to find the gradient of the error function $E$ with respect to $W_{jl}$. The gradient of the error function with respect to the $W_{jlR}$ and $W_{jlI}$ are given by
\begin{eqnarray}
\frac{\partial E}{\partial W_{jlR}} & = & \frac{\partial E}{\partial u^{j}} \bigg ( \frac{\partial u^{j}}{\partial x_{j}} \frac{\partial x_{j}}{\partial W_{jlR}} + \frac{\partial u^{j}}{\partial y_{j}} \frac{\partial y_{j}}{\partial W_{jlR}} \bigg ) \nonumber \\
&+& \frac{\partial E}{\partial v^{j}} \bigg ( \frac{\partial v^{j}}{\partial x_{j}} \frac{\partial x_{j}}{\partial W_{jlR}} + \frac{\partial v^{j}}{\partial y_{j}} \frac{\partial y_{j}}{\partial W_{jlR}} \bigg ) \\
 &=& -\delta_{JR}(u^{j}_{x}X_{jlR} + u^{j}_{y}X_{jlI})  \nonumber \\
 & & - \delta_{JI}(v^{j}_{x}X_{jlI} + v^{j}_{y}X_{jlI}) 
\label{eq: E_partial_WR}
\end{eqnarray}
\begin{eqnarray}
\frac{\partial E}{\partial W_{jlI}} &=& \frac{\partial E}{\partial u^{j}} \bigg ( \frac{\partial u^{j}}{\partial x_{j}} \frac{\partial x_{j}}{\partial W_{jlI}} + \frac{\partial u^{j}}{\partial y_{j}} \frac{\partial y_{j}}{\partial W_{jlI}} \bigg ) \nonumber \\
&+& \frac{\partial E}{\partial v^{j}} \bigg ( \frac{\partial v^{j}}{\partial x_{j}} \frac{\partial x_{j}}{\partial W_{jlI}} + \frac{\partial v^{j}}{\partial y_{j}} \frac{\partial y_{j}}{\partial W_{jlI}} \bigg )\\
&=& -\delta_{JR}(u^{j}_{x}(-X_{jlI}) + u^{j}_{y}(X_{jlR}) \nonumber \\
& & - \delta_{JI}(v^{j}_{x}(-X_{jlI}) + v^{j}_{y}X_{jlR}) 
\label{eq: E_partial_WI}
\end{eqnarray}
where $\delta_{j} = -\partial E / \partial u^{j} - i\partial E / \partial v^{j}$, $\delta_{jR} = -\partial E / \partial u^{j}$ and $\delta_{jI} = -\partial E / \partial v^{j}$.  By combining equations (\ref{eq: E_partial_WR}) and (\ref{eq: E_partial_WI}), the gradient of the error function with respect to $W_{jl}$ is given by
\begin{eqnarray}\nonumber
\nabla_{wjl}E &=&  \frac{\partial E}{\partial W_{jlR}} + i\frac{\partial E}{\partial W_{jlI}}\\
&=& -\bar{X}_{jl}((u_{x}^{j} + iu_{y}^{j})\delta_{jR} + (v_{x}^{j} + iv_{y}^{j})\delta_{jI})
\label{eq:gradient_E}
\end{eqnarray}
Hence, given a positive constant learning rate $\alpha$, the complex weight $W_{jl}$ must be changed by a value $\Delta W_{jI}$ proportional to the negative gradient:
\begin{equation}
\Delta W_{jl} = \alpha \bar{X}_{jl} \left( (u_{x}^{j} + iu_{y}^{j})\delta_{jR} + (v_{x}^{j} + iv_{y}^{j})\delta_{jI} \right)
\label{eq:delta_W}
\end{equation}
For an output neuron, $\delta_{jR}$ and $\delta_{jI}$ in equation (\ref{eq:gradient_E}) are given by
\begin{eqnarray}\nonumber
\delta_{jR} = \frac{\delta E}{\delta u^{j}} = \epsilon_{jR} = d_{jR} - u^{j} \\
\delta_{jI} = \frac{\delta E}{\delta v^{j}} = \epsilon_{jI} = d_{jI} - v^{j}
\label{eq:expanded_output_error}
\end{eqnarray}
And in compact form, it will be
\begin{equation}
\delta_{j} = \epsilon_{j} = d_{j} - o_{j}
\label{eq:output_error}
\end{equation}
The chain rule is used to compute $\delta_{jR}$ and $\delta_{jI}$ for the hidden neuron. Note that $k$ is an index for a neuron receiving input from neuron $j$. The net input $z_{k}$ to neuron $k$ is
\begin{equation}
z_{k} = x_{k} + iy_{k} = \sum_{l}(u^{l} + iv^{l})(W_{klR} + iW_{klI})
\end{equation}
where $l$ is the index for the neurons that feed into neuron $k$. 
Computing $\delta_{jR}$ using the chain rule yields
\begin{eqnarray}\nonumber
\delta_{jR} = - \frac{\delta E}{\delta u^{j}} &=& - \sum_{k} \frac{\delta E}{\delta u^{k}} \bigg (\frac{\delta u^{k}}{\delta x_{k}} \frac{\delta x_{k}}{\delta u^{j}} + \frac{\delta u^{k}}{\delta y_{k}} \frac{\delta y_{k}}{\delta u^{j}}\bigg )\\\nonumber
& & - \sum_{k} \frac{\delta E}{\delta v^{k}} \bigg (\frac{\delta v^{k}}{\delta x_{k}} \frac{\delta x_{k}}{\delta u^{j}} + \frac{\delta v^{k}}{\delta y_{k}} \frac{\delta y_{k}}{\delta u^{j}} \bigg )\\\nonumber
& = & \sum_{k}\delta_{kR}(u^{k}_{x}W_{kjR}+ u^{k}_{y}W_{kjI})\\
& & + \sum_{k}\delta_{kI}(v^{k}_{x}W_{kjR}+ v^{k}_{y}W_{kjI})
\label{eq:hidden_weight_updateR}
\end{eqnarray}
Similarly, $\delta_{jI}$ is computed as:
\begin{eqnarray}\nonumber
\delta_{jI} = - \frac{\delta E}{\delta v^{j}} = - \sum_{k} \frac{\delta E}{\delta u^{k}} \bigg (\frac{\delta u^{k}}{\delta x_{k}} \frac{\delta x_{k}}{\delta v^{j}} + \frac{\delta u^{k}}{\delta y_{k}} \frac{\delta y_{k}}{\delta v^{j}}\bigg )\\\nonumber
- \sum_{k} \frac{\delta E}{\delta v^{k}} \bigg (\frac{\delta v^{k}}{\delta x_{k}} \frac{\delta x_{k}}{\delta v^{j}} + \frac{\delta v^{k}}{\delta y_{k}} \frac{\delta y_{k}}{\delta v^{j}} \bigg )\\\nonumber
= \sum_{k}\delta_{kR}(u^{k}_{x}(-W_{kjI})+ u^{k}_{y}W_{kjR})\\
+ \sum_{k}\delta_{kI}(v^{k}_{x}W_{kjR}+ v^{k}_{y}W_{kjI})
\label{eq:hidden_weight_updateI}
\end{eqnarray}
The expression for $\delta_{j}$ is obtained by combining equations (\ref{eq:hidden_weight_updateR}) and (\ref{eq:hidden_weight_updateI}):
\begin{eqnarray}\nonumber
\delta_{j} &=& \delta_{jR} + i\delta_{jI} \\
&=& \sum_{k} \bar{W}_{kj} \left( (u^{k}_{x} + iu^{k}_{y})\delta_{kR} + (v^{k}_{x} +iv^{k}_{y})\delta_{kl} \right)
\label{eq:update_rule}
\end{eqnarray}
where $\delta_{j}$ is computed for neuron $j$ starting in the output layer using equation (\ref{eq:output_error}), then using equation (\ref{eq:update_rule}) for the neurons in the hidden layers. After computing $\delta_{j}$ for neuron $j$, equation (\ref{eq:delta_W}) is used to update its weights. 

\subsection{Non-Gradient-based Approach}
Different from the gradient based approach, the learning process of a neural network based on the multi-valued neuron (MVN) is derivative-free and it is based on the error-correction learning rule~\cite{Aizenberg1996}. 
For a single neuron, weight correction in the MVN is determined by the neuron's error, and learning is reduced to a simple movement along the unit circle. The corresponding activation function of equations (\ref{eq:discAF}) and (\ref{eq:contAF}) are not differentiable, which implies that there are no gradients.

\begin{figure}
	 \centering
    	 \includegraphics[width=0.4\textwidth]{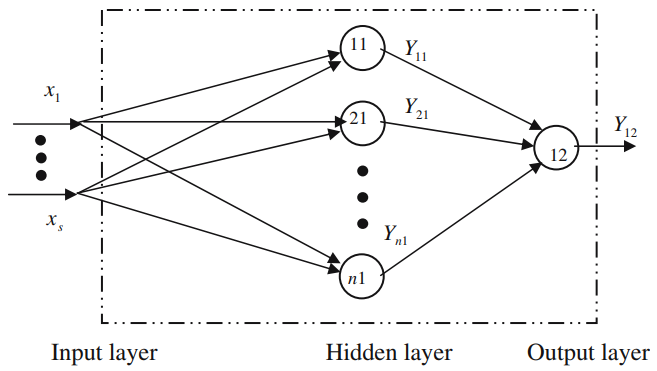}
      	\caption{Example of MLMVN with one hidden-layer and a single output}
    \label{fig:MLMVN}
\end{figure}

Considering an example MLMVN with one hidden-layer and a single output as shown in Figure~\ref{fig:MLMVN}. If $T$ is the target, $Y_{12}$ is the output, and the following definitions are assumed: 
\begin{itemize}
\item $\epsilon^{*} = T - Y_{12}$: global error of network
\item $w^{12}_{0}, w^{12}_{1},\cdots, w^{12}_{n}$: initial weighting vector of neuron $Y_{12}$
\item $Y_{i1}$: initial output of neuron $Y_{12}$
\item $Z_{12}$: weighed sum of neuron $Y_{12}$ before weight correction
\item $\epsilon_{12}$: error of neuron $Y_{12}$
\end{itemize}

The weight correction for the second to the $m$th (output) layer, and then for the input layer are given by
\begin{eqnarray}\nonumber
\tilde{w}^{kj}_{i} &=& w^{kj}_{i} + \frac{C_{kj}}{(N_{j-1}+1)}\epsilon_{kj}\bar{\tilde{Y}}_{i,j-1}, \quad \text{$i=1, \dots ,n$}\\
\tilde{w}^{kj}_{0} &=& w^{kj}_{0} + \frac{C_{kj}}{(N_{j-1}+1)}\epsilon_{kj}
\label{eq:output_correction_rule}
\end{eqnarray}
\begin{eqnarray}\nonumber
\tilde{w}^{k1}_{i} &=& w^{k1}_{i} + \frac{C_{k1}}{(n+1)}\epsilon_{k1}\bar{x_{i}}, \quad \text{$i=1, \dots ,n$}\\
\tilde{w}^{k1}_{0} &=& w^{k1}_{0} + \frac{C_{k1}}{(n+1)}\epsilon_{kj}
\label{eq:input_correction_rule}
\end{eqnarray}
$C_{kj}$ is the learning rate for the $k$th neuron of the $j$th layer. However, in applying this learning rule  two situations may arise: (1) the absolute value of the weighted sum being corrected may jump erratically, or (2) the output of the hidden neuron varies around some constant value. In either of these scenarios, a large number of weight updates can be wasted. The workaround is to instead apply a modified learning rule~\cite{Aizenberg2014a} which adds a normalization constant to the learning for the hidden and input layers. However, the output layer learning error back propagation is not normalized. The final correction rule for the $k$th neuron of the $m$th (output) layer is
\begin{eqnarray}\nonumber
\tilde{w}^{km}_{i} &=& w^{km}_{i} + \frac{C_{km}}{(N_{m-1}+1)}\epsilon_{km}\bar{\tilde{Y}}_{i,m-1}, \quad \text{$1=1, \dots ,n$}\\
\tilde{w}^{km}_{0} &=& w^{km}_{0} + \frac{C_{km}}{(N_{m-1}+1)}\epsilon_{km}
\label{eq:final_output_correction_rule}
\end{eqnarray}
For the second till the $(m-1)$th layer ($k$th neuron of the $j$th layer ($j = 2, \cdots , m-1$), the correction rule is
\begin{eqnarray}\nonumber
\tilde{w}^{kj}_{i} &=& w^{kj}_{i} + \frac{C_{kj}}{(N_{j-1}+1)}\epsilon_{kj}\bar{\tilde{Y}}_{i,j-1}, \quad \text{$1=1, \dots ,n$}\\
\tilde{w}^{kj}_{0} &=& w^{kj}_{0} + \frac{C_{kj}}{(N_{j-1}+1) \vert z_{kj} \vert}\epsilon_{kj}
\label{eq:final_hiddens_correction_rule}
\end{eqnarray}
and for the input layer:
\begin{eqnarray}\nonumber
\tilde{w}^{k1}_{i} &=& w^{k1}_{i} + \frac{C_{k1}}{(n+1)}\epsilon_{k1}\bar{x_{i}}, \quad \text{$1=1, \dots ,n$}\\
\tilde{w}^{k1}_{0} &=& w^{k1}_{0} + \frac{C_{k1}}{(n+1)  \vert z_{kj} \vert}\epsilon_{kj}
\label{eq:final_input_correction_rule}
\end{eqnarray}
Given a pre-specified learning precision $\omega$, the condition for termination of the learning process is
\begin{equation}
\frac{1}{N} \sum_{s=1}^{N} \sum_{k} (\epsilon^{*}_{km_{s}})^{2}(W) = \frac{1}{N} \sum_{s=1}^{N} E_{s} \leq \omega .
\end{equation}
One of the advantages of the non-gradient based approach is its ease of implementation. In addition, since there are no derivatives involved, the problems associated with typical gradient descent based methods will not apply here, such as problem with being stuck in a local minima. Furthermore, because of the structure and representation of the network, it is possible to design a hybrid network architecture where some nodes have discrete activation functions, whereas some others use continuous activation functions. This may have a great potential in future applications~\cite{Aizenberg2007}.

\subsection{Training and hyperparameters optimization }
\label{subsec:hyperparameters}
The adaptation of real-valued activation, weight initialization and batch normalization was analyzed in~\cite{Trabelsi2017}.
More recently, building on the work of~\cite{Brandwood1983}, optimization via second-order methods were introduced by the use of the complex Hessian in the complex gradient~\cite{van1994}, and  linear approaches have been proposed to model second-order complex statistics~\cite{Goh2008}. The training of complex-valued neural networks using complex learning rate was proposed in~\cite{Zhang2016}. The problem of vanishing gradient in recurrent neural networks was solved by using unitary weight matrices in~\cite{Arjovsky2016} and~\cite{Wisdom2016}, thereby improving time and space efficiency by exploiting the properties of orthogonal matrices. However, regarding regularization, besides the proposed use of noise in~\cite{Hirose1999}, not much work has been done in the literature. This is an interesting open research problem and it is further discussed in Section~\ref{sec:challenges}.

\section{Input and Output representations in CVNNs}
\label{sec:in_out}
Input representations can be  complex either naturally or by design. In the former case, a set of complex numbers represent the data domain. An example is Fourier transform on images. In the latter case, inputs have magnitude and phase which are statistically correlated. An example is in radio frequency or  wind data~\cite{Mandic2008}. For the outputs, complex values are more suitable for regression tasks. However, real-values are more appropriate if the goal is to perform inference over a probability distribution of complex parameters. Weights can be complex or real  irrespective of  input/output representations. 

In~\cite{Saroff2018}, the impact and tradeoff on choices that may be made on representations was studied using a toy experiment. In the experiment, a toy model was trained to learn a function that is able to add sinusiods. Four input representations were used:
\begin{enumerate} 
\item amplitude-phase where a real-valued vector is formed from concatenating the phase offset and amplitude parameters; 
\item complex representation where the phase offset is used as the phase of the complex phasor; 
\item real-imaginary representation where the real and imaginary components of the complex vector in 2) are used as real vectors;
\item augmented complex representation which is a concatenation of the complex vector and its conjugate.
\end{enumerate}
Two representations were considered for the target: 
\begin{enumerate} 
\item  the straightforward representation which use the raw real and complex target values;
\item  analytic representation which uses the Hilbert transform for the imaginary part. 
\end{enumerate}
The activation functions used were (1) identity, (2) hyperbolic tangent, (3) split real-imaginary activation, and (4) split amplitude phase. 

Different models with the combinations of input and output representations as well as various activation functions were tested in the experiments, and the model with the real-imaginary sigmoid activation performed the best. However,  it is interesting that  this best model diverged and performed poorly when it was trained on real inputs. There is a closed form for the real-imaginary input representation when the output target representation are either straightforward or analytic. However, there is no closed form solution for the amplitude-phase representation. 

In general, certain input and output representations when combined yield a closed form solution, and there are some scenarios in which even when there is no closed form solution, the complexity of the solution can be reduced by transforming either the input, output or internal representation. In summary, the question as to which input and output representation is the best would typically depend on the application and it is affected by the level of constraint imposed on the system.

\section{Applications of CVNNs}
\label{sec:applications}
Various applications of CVNNs are summarized in Table~\ref{table:applicationsCVNN}. Because of the complex nature of many natural and man-made signals, CVNNs find most of their applications in the area of signal (including radio frequency signal, audio and image) processing.
\subsection{Applications in Radio Frequency Signal Processing in Wireless Communications}
\label{subsec:applicationsSignalProcessing}
The majority of the work on complex-valued neural networks has been focused on signal processing research and applications. Complex-valued neural network research in signal processing applications include channel equalization~\cite{Chen2000,Solazzi2002}, satellite communication equalization~\cite{Benvenuto1991}, adaptive beamforming~\cite{Suksmono2003}, coherent-lightwave networks~\cite{HIROSE1994,Hirose2000} and source separation~\cite{Scarpiniti2008}. In interferometric synthetic aperture radar, complex networks were used for adaptive noise reduction~\cite{Oyama2018}. In electrical power systems, complex valued networks were proposed to enhance power transformer modeling~\cite{Chistyakov2011}, and analysis of load flow~\cite{Ceylan2005}. In~\cite{Benvenuto1991} for example, the authors considered the issue of requiring long sequences for training, which results in a lot of wasted channel capacity in cases where nonlinearities associated with the channel are slowly time varying. To solve this problem, they investigated the use of CVNN for adaptive channel equalization. The approach was tested on the task of equalizing a digital satellite radio channel amidst intersymbol interference and minor nonlinearities, and their approach showed competitive results. They also pointed out that their approach does not require prior knowledge about the nonlinear characteristics of the channel.

\begin{table}[ht!]
\caption{Applications of Complex-Valued Neural Networks}
\begin{tabular}{|p{3cm}|p{4.6cm}|}
 \hline
{\bf Applications} & {\bf Corresponding Publications} \\ 
\hhline{|=|=|}
Radio Frequency Signal Processing in Wireless Communications  & \cite{Widrow1960,Widrow1975,Adali2000,Kim2002,Inhyok1995,Chen1994a,Chen1994b,Deng2002,Uncini1999,Scarpiniti2008,Scardapane2018,Gong2017,Zhang2017b,Liu2017,Peker2016,Hirose2012,Hu2018a,Hu2018b,Ding2014,Suzuki2013,Yuan2019,Aizenberg2016a,Catelani2016,Marseet2017,Chen2000,Solazzi2002,Benvenuto1991,Suksmono2003,Hirose2000,Chistyakov2011,Chistyakov2012}\\\hline

Image Processing and Computer Vision  & \cite{Zemel1995, Aizenberg1996,Arjovsky2016, Trabelsi2017,Virtue2017,Gleich2018,Popa2017a,Amilia2015,Liu2014,Olanrewaju2011,Hata2016,Kominami2017,Popa2018a,Li2019,Gu2018,Matlacz2018,Popa2018b,Aizenberg2018a,Aizenberg2014a,Aizenberg2011a,Aizenberg2008b,Aizenberg2000,Aizenberg2000b,Miyauchi1993,Miyauchi1992,Hirose2006b,Hirose2006a,Aizenberg2008,Aizenberg2000c}  \\\hline 

Audio Signal Processing and Analysis & \cite{Trabelsi2017,Hayakawa2018,Kataoka1998,Kinouchi1996,Al_Nuaimi2012,
Lee2017,Tsuzuki2013}  \\\hline 

Radar / Sonar Signal Processing & \cite{Aizenberg2007,Gao2019,Wilmanski2016,Yao2019,Yao2020,Oyama2018} \\\hline

Cryptography & \cite{Dong2019}  \\\hline

Time Series Prediction & \cite{Aizenberg2007,Aizenberg1996}  \\\hline

Associative Memory & \cite{Jankowski1996,Miyajima2000} \\\hline

Wind Prediction & \cite{Cevik2018,Kitajima2010,Mandic2008}  \\\hline

Robotics  & \cite{Maeda2014}  \\\hline

Traffic Signal Control (robotics)  & \cite{Nishikawa2005,Nishikawa2006}  \\\hline

Spam Detection  & \cite{Hu2008}  \\\hline

Precision Agriculture (soil moisture prediction) & \cite{Aizenberg2018a}  \\
\hhline{|=|=|}
\end{tabular}
\label{table:applicationsCVNN}
\end{table}

\subsection{Applications in Image Processing and Computer Vision}
\label{subsec:applicationsComputerVision}
Complex valued neural networks have also been applied in image processing and computer vision. There have been some works on applying CVNN for optical flow~\cite{Miyauchi1992,Miyauchi1993}, and CVNNs have been combined with holographic movies~\cite{Hirose2006a,Hirose2006b,Tay2007}. CVNNs were used for reconstruction of gray-scale images~\cite{Tanaka2009}, and image deblurring~\cite{Aizenberg2008,Aizenberg2008b,Aizenberg2000c}, classification of microarray gene expression~\cite{Aizenberg2006}.  Clifford networks were applied for character recognition~\cite{Rahman2001} and complex-valued neural networks were also applied for automatic gender recognition in~\cite{Amilia2015}. A complex valued VGG network was implemented by~\cite{Gu2018} for image classification. In this work, building on~\cite{Trabelsi2017}, the building blocks of the VGG network including batch normalization, ReLU activation function, and the 2-D convolution operation were transformed to the complex domain. When testing their model on classification of the popular CIFAR10 benchmark image dataset, the complex-valued VGG model performed slightly better than the real-valued VGG in both training and testing accuracy. Moreover, the complex-valued network requires less parameters.

Another noteworthy computer vision application that showcases the potential of complex-valued neural networks can be found in~\cite{Arjovsky2016}, where the authors investigated the benefits of using unitary weight matrices to mitigate the vanishing gradient problem. Among other applications, their method was tested on the MNIST hand-writing benchmark dataset in two modes. In the first mode, the pixels were read in order (left to right and bottom up), and in the second mode the pixels were read in arbitrarily. The real-valued LSTM performed slightly better than the unitary-RNN for the first mode, but in the second mode the unitary-RNN outperformed the real-valued LSTM in spite of having below a quarter of the parameters than the real-valued LSTM. Moreover, it took the real-valued LSTM between 5 and 10 times as many epochs to reach convergence comparing to the unitary RNN.

Non-gradient based learning has also been used extensively in image processing and computer vision applications. For example, the complex neural network with multi-valued neurons (MLMVN) was used as an intelligent image filter in~\cite{Aizenberg2017a}. The task was to apply the filter to simultaneously filter all pixels in an $n$ x $n$ region, and the results from overlapping regions of paths were then averaged. The integer input intensities were mapped to complex inputs before being fed into the MLMVN, and the integer output from the MLMVN were transformed to integer intensities. This approach proved successful and efficient, as very good nonlinear filters were obtained by training the network with as little as 400 images. Furthermore, from their simulations it was observed that the filtering results got better as more images were added to the training set.

\subsection{Applications in Audio Signal Processing and Analysis}
\label{subsec:applicationsAudioProcessing}
In audio processing, complex networks were proposed to improve the MP3 codec in~\cite{Al_Nuaimi2012}, and in audio source localization~\cite{Tsuzuki2013}. CVNNs were shown to denoise noise-corrupted waveforms better than real-valued neural  networks in~\cite{Hirose2012}. Complex associative memories were proposed for temporal series obtained from symbolic music representation~\cite{Kataoka1998,Kinouchi1996}. A notable work in this regard is~\cite{Trabelsi2017} where deep complex networks were used for speech spectrum prediction and music transcription. In this work, the authors compared the various complex-valued ReLU-based activations and formulated the building blocks necessary for building deep complex networks. The building blocks include a complex batch normalization, weight initialization. Apart from image recognition, the authors tested their complex deep network on MusicNet dataset for the music transcription task, and on the TIMIT dataset for the Speech Spectrum prediction tasks. It is interesting to note that the datasets used contain real values, and the imaginary components are learned using operations in one real-valued residual block.

\subsection{Other Applications }
\label{subsec:Otherapplications}
In wind prediction, the axes of the Cartesian coordinates of a complex number plane was used to represent the cardinal points (north, south, east, west), and the prediction of wind strength was expressed using the distance from the origin in~\cite{Popovic2004}. However, although the circularity property of complex numbers were exploited in this work, this method does not reduce the degree of freedom, instead it has the same degree of freedom as a real-valued network because of the typically high anisotropy of wind. In other words, in this representation, the absolute value of phase does not yield any meaning. Rather, it is the difference from a specified reference that is meaningful.

A recent application in radar can be found in~\cite{Yao2019,Yao2020}. In this work, a complex-valued convolutional neural network (CV-CNN) was used to exploit the inherent nature of the time-frequency data of human echoes to classify human activities. The human echo is a combination of the phase modulation information caused by motion, and the amplitude data obtained from different parts of the body. Short Time Fourier Transform was used to transform the human radar echo signals and used to train the CV-CNN. The authors also demonstrated that their method performed better than other machine learning approaches at low signal-to-noise ratio (SNR), while achieving accuracies as high as 99.81$\%$. CV-CNNs were also used in~\cite{Tan2020} to enhance radar images. Interestingly, this is an application of CNN to a regression problem, where the inputs are radar echoes and the outputs are expected images. 

The first use of a complex-valued generative adversarial network (GAN) was proposed in~\cite{Sun2019} to mitigate the issue of lack of labeled data in polarimetric synthetic aperture radar (PolSAR) images. This approach which retains the amplitude and phase information of the PolSAR images performed better than state-of-the-art real-valued approaches, especially in scenarios involving fewer annotated data.

A complex-valued tree parity machine network (CVTPM) was used for neural cryptography in~\cite{Dong2019}. In neural cryptography, by applying the principle of neural network synchronization, two neural networks exchange public keys. The benefit of using a complex network over a real network is that two group keys can be exchanged in one process of neural synchronization. Furthermore, compared to a real network with the same architecture, the CVTPM was shown to be more secure.

There have been other numerous applications of CVNNs. For instance,  a discriminative complex-valued convolutional neural network was applied to electroencephalography (ECG) signals in~\cite{Zhang2017b} to automatically deduce features from the data and predict stages of sleep. By leveraging the Fisher criterion which takes into account both the minimum error, the maximum between-class distance, and the minimum within-class distance, the authors assert that their model named  fast discriminative complex-valued convolutional neural network (FDCCNN) is capable of learning inherent features that are discrimitative enough, even when the dataset is imbalanced.

\section{Challenges and Potential Research}
\label{sec:challenges}
Complex valued neural networks have been shown to have potential in domains where representation of the data encountered is naturally complex, or complex by design. Since the 1980s, research of both real valued neural networks (RVNN) and complex valued neural networks (CVNN)  have advanced rapidly. However, during the development of deep learning,  research on CVNNs has not been very active compared to RVNNs. So far research on CVNNs has mainly targeted at shallow architectures, and specific signal processing applications such as channel equalization. One reason for this is the difficulty associated with training. \emph{This is due to the limitation that the complex-valued activation is not complex-differentiable and bounded at the same time}. Several studies~\cite{Trabelsi2017} have suggested that the constraint of requiring that a complex-valued activation be simultaneously bounded and complex differentiable need not be met, and propose activations that are differentiable independently with respect to the real and imaginary components. This remains an open area of research.

Another reason for the  slow development of research in complex-valued neural networks is that almost all deep learning libraries are optimized for real-valued operations and networks. There are hardly any public libraries developed and optimized specifically for training CVNNs. In the experiments performed in~\cite{Anderson2017} and validated in~\cite{Joshua2019}, a baseline, wide and deep network was built for real, complex and split valued neural networks. Computations were represented by computational sub-graphs of operations between the real and imaginary parts. This enabled the use of TensorFlow, a standard neural network libraries instead of the generalized derivatives in Clifford algebra. However, this approach to modeling CVNN is still fundamentally based on a library meant for optimized real-valued arithmetic computations. This issue concerns practical implementation, and \emph{there is a need for deep learning libraries targeted and optimized  for complex-valued computations}. 

Regarding weight initialization,  a formulation for complex weight initialization was presented in~\cite{Trabelsi2017}. By formulating this in terms of the variance of the magnitude of the weights which follows the Rayleigh distribution with two degrees of freedom, the authors represented the variance of the weights in terms of the Rayleigh distribution's parameter. It is shown that the variance of the weights depends on the magnitude and not on the phase, hence the phase was initialized uniformly within the range $[-\pi, \pi]$.  More research on this could provide  insights and  yield meaningful results as to \emph{alternative methods of complex-valued weight initialization for CVNNs}. For instance, weight initialization scheme could make use of the phase parameter in addition to the magnitude.

There has been some strides made in applications involving complex valued recurrent networks. For example, the introduction of unitary matrices which are the generalized complex version of real-valued Hermitian matrices, mitigates the problem of vanishing and exploding gradients. However,  in applications involving long sequences, gradient backpropagation requires all hidden states values to be stored. This can become impractical given the limited availability of GPU memory for optimization. Considering that the inverse of the unitary matrix is its conjugate transpose,  it may be possible to derive some invertible nonlinear function with which the states can be computed during the backward pass. This will eliminate the need to store the hidden state values.

By introducing complex parameters, the number of operations required increases thus increasing the computational complexity. Compared to real-valued parameters which use single real-valued multiplication, complex-valued parameters will require up to four real multiplications and two real additions. This means that merely doubling the number of real-valued parameters in each layer does not give the equivalent effect that is observed in a complex-valued neural network~\cite{Nils2018}. Furthermore, the capacity of a network in terms of its ability to approximate structurally complex functions can be quantified by the number of (real-valued) parameters in the network. Consequently, by representing a complex number $a + ib$ using real numbers $(a,b)$, the number of real parameters for each layer is doubled. This implies that by using a complex-valued network, a network may gain more expressiveness but run the risk of overfitting due to the increase in parameters as the network goes deeper. Hence, \emph{regularization during the training of CVNNs is important but remains an open problem}. 

Ridge ($L_{2}$) and LASSO ($L_{1}$) regularizations are the two forms of regression that are aimed at mitigating the effects of multicollinearity. Regularization enforces an upper threshold on the values of the coefficients and produces a solution with coefficients of smaller variance. 
With the $L_{2}$ regularization, the update to the weights is penalized. This forces the magnitude of the weights to be small and hence reduce overfitting. Depending on the application, $L_{1}$ norm can be applied that will force some weights to be zero (sparse representation).

However, the issue with the above formulation is that it cannot be directly applied over the field of complex numbers. This is because the  application of the above formulation of the $L_{2}$ norm to a complex number is simply its magnitude, which is not a complex number and the phase information is lost. As far as we know, there has not been any successful attempt to either provide a satisfactory transformation of this problem into the complex domain, or derive a method that is able to efficiently search the underlying hyper-parameter solution space. Apart from the work by the authors in~\cite{Hirose1999} who propose to use noise, there is very little work on the regularization of CVNNs. 

Complex and split-complex-valued neural networks are considered in~\cite{Anderson2017} to further understand their computational graph, algebraic system and expressiveness. This result shows that the complex-valued neural network are more sensitive to hyperparameter tuning due to the increased complexity of the computational graph. In one of the experiments performed in~\cite{Anderson2017}, $L_{2}$ regularization was added for all parameters in the real, complex and split-complex neural networks and trained on MNIST and CIFAR-10 benchmark datasets. It was observed that in the un-regularized case, both real and complex networks showed comparable validation accuracy. However, when $L_{2}$ regularization was added, overfitting was reduced in the real-valued networks but had very little effect on the performance of the complex and split-complex networks. It seems that complex neural networks are not self regularizing, and they are more difficult to regularize than their real-valued counterpart. 

\section{Conclusion}
\label{sec:conclusion}
A comprehensive review on complex-valued neural networks has been presented in this work. The argument for advocating the use of complex-valued neural networks in domains where complex numbers occur naturally or by design was presented. The state-of-the-art in complex-valued neural networks was presented by classifying them according to activation function, learning paradigm, input and output representations, and applications. Open problems and future research directions have also been discussed. Complex-valued neural networks compared to their real-valued counterparts are still considered an emerging field and require more attention and actions from the deep learning and signal processing research community.

\section{Acknowledgment}
\label{sec:acknowledgement}
This research work is supported by the U.S. Office of the Under Secretary of Defense for Research and Engineering (OUSD(R\&E)) under agreement number FA8750-15-2-0119. The U.S. Government is authorized to reproduce and distribute reprints for governmental purposes notwithstanding any copyright notation thereon. The views and conclusions contained herein are those of the authors and should not be interpreted as necessarily representing the official policies or endorsements, either expressed or implied, of the Office of the Under Secretary of Defense for Research and Engineering (OUSD(R\&E)) or the U.S. Government.

\bibliographystyle{IEEEtran}
\bibliography{MLMVN-01-17-2021}

\begin{thebibliography}{100}

\bibitem{Saroff2018}
A.~M. Sarroff, ``{Complex Neural Networks for Audio},'' Tech. Rep. TR2018-859,
  Dartmouth College, Computer Science, Hanover, NH, May 2018.

\bibitem{Shinde2018}
P.~P. {Shinde} and S.~{Shah}, ``A review of machine learning and deep learning
  applications,'' in {\em 2018 Fourth International Conference on Computing
  Communication Control and Automation (ICCUBEA)}, pp.~1--6, 2018.

\bibitem{Hirose2011}
A.~Hirose and S.~Yoshida, ``Comparison of complex- and real-valued feedforward
  neural networks in their generalization ability,'' in {\em Neural Information
  Processing - 18th International Conference, {ICONIP}, 2011, Shanghai, China,
  November 13-17, 2011, Proceedings, Part {I}}, pp.~526--531, 2011.

\bibitem{HIROSE1994}
A.~Hirose, ``Applications of complex-valued neural networks to coherent optical
  computing using phase-sensitive detection scheme,'' {\em Information Sciences
  - Applications}, vol.~2, no.~2, pp.~103 -- 117, 1994.

\bibitem{Hirose1992}
A.~{Hirose}, ``Continuous complex-valued back-propagation learning,'' {\em
  Electronics Letters}, vol.~28, pp.~1854--1855, Sep. 1992.

\bibitem{Widrow1960}
B.~Widrow and M.~E. Hoff, ``Adaptive switching circuits,'' in {\em 1960 {IRE}
  {WESCON} Convention Record, Part 4}, (New York), pp.~96--104, {IRE}, 1960.

\bibitem{Rosenblatt1958}
F.~F. Rosenblatt, ``The perceptron: a probabilistic model for information
  storage and organization in the brain.,'' {\em Psychological review}, vol.~65
  6, pp.~386--408, 1958.

\bibitem{Widrow1975}
B.~{Widrow}, J.~{McCool}, and M.~{Ball}, ``The complex lms algorithm,'' {\em
  Proceedings of the IEEE}, vol.~63, pp.~719--720, April 1975.

\bibitem{Brandwood1983}
D.~H. {Brandwood}, ``A complex gradient operator and its application in
  adaptive array theory,'' {\em IEE Proceedings H - Microwaves, Optics and
  Antennas}, vol.~130, pp.~11--16, February 1983.

\bibitem{Wirtinger1927}
W.~Wirtinger, ``Zur formalen theorie der funktionen von mehr komplexen
  ver{\"a}nderlichen,'' {\em Mathematische Annalen}, vol.~97, pp.~357--375, Dec
  1927.

\bibitem{Hirose2012}
A.~{Hirose} and S.~{Yoshida}, ``Generalization characteristics of
  complex-valued feedforward neural networks in relation to signal coherence,''
  {\em IEEE Transactions on Neural Networks and Learning Systems}, vol.~23,
  pp.~541--551, April 2012.

\bibitem{Reichert2014}
D.~P. Reichert and T.~Serre, ``Neuronal synchrony in complex-valued deep
  networks,'' {\em CoRR}, vol.~abs/1312.6115, 2014.

\bibitem{Srivastava2015}
R.~K. Srivastava, K.~Greff, and J.~Schmidhuber, ``Training very deep
  networks,'' in {\em Advances in Neural Information Processing Systems 28}
  (C.~Cortes, N.~D. Lawrence, D.~D. Lee, M.~Sugiyama, and R.~Garnett, eds.),
  pp.~2377--2385, Curran Associates, Inc., 2015.

\bibitem{Cho2014}
K.~Cho, B.~van Merrienboer, D.~Bahdanau, and Y.~Bengio, ``On the properties of
  neural machine translation: Encoder-decoder approaches,'' in {\em
  SSST@EMNLP}, 2014.

\bibitem{Shi2006}
G.~Shi, M.~M. Shanechi, and P.~Aarabi, ``On the importance of phase in human
  speech recognition,'' {\em Trans. Audio, Speech and Lang. Proc.}, vol.~14,
  p.~1867–1874, Sept. 2006.

\bibitem{Oppenheim1981}
A.~V. {Oppenheim} and J.~S. {Lim}, ``The importance of phase in signals,'' {\em
  Proceedings of the IEEE}, vol.~69, no.~5, pp.~529--541, 1981.

\bibitem{Danihelka2016}
I.~Danihelka, G.~Wayne, B.~Uria, N.~Kalchbrenner, and A.~Graves, ``Associative
  long short-term memory,'' in {\em Proceedings of the 33rd International
  Conference on International Conference on Machine Learning - Volume 48},
  ICML'16, pp.~1986--1994, JMLR.org, 2016.

\bibitem{He2016}
K.~{He}, X.~{Zhang}, S.~{Ren}, and J.~{Sun}, ``Deep residual learning for image
  recognition,'' in {\em 2016 IEEE Conference on Computer Vision and Pattern
  Recognition (CVPR)}, pp.~770--778, 2016.

\bibitem{Arjovsky2016}
M.~Arjovsky, A.~Shah, and Y.~Bengio, ``Unitary evolution recurrent neural
  networks,'' in {\em Proceedings of the 33rd International Conference on
  International Conference on Machine Learning - Volume 48}, ICML'16,
  pp.~1120--1128, JMLR.org, 2016.

\bibitem{Wisdom2016}
S.~Wisdom, T.~Powers, J.~R. Hershey, J.~L. Roux, and L.~Atlas, ``Full-capacity
  unitary recurrent neural networks,'' in {\em Proceedings of the 30th
  International Conference on Neural Information Processing Systems}, NIPS'16,
  (USA), pp.~4887--4895, Curran Associates Inc., 2016.

\bibitem{Cybenko1989}
G.~Cybenko, ``Approximation by superpositions of a sigmoidal function,'' 1989.

\bibitem{Barron1994}
A.~Barron, ``Approximation and estimation bounds for artificial neural
  networks,'' vol.~14, pp.~243--249, 01 1991.

\bibitem{FUNAHASHI1989}
K.-I. Funahashi, ``On the approximate realization of continuous mappings by
  neural networks,'' {\em Neural Networks}, vol.~2, no.~3, pp.~183 -- 192,
  1989.

\bibitem{HORNIK1989}
K.~Hornik, M.~Stinchcombe, and H.~White, ``Multilayer feedforward networks are
  universal approximators,'' {\em Neural Networks}, vol.~2, no.~5, pp.~359 --
  366, 1989.

\bibitem{Benvenuto1992}
N.~{Benvenuto} and F.~{Piazza}, ``On the complex backpropagation algorithm,''
  {\em IEEE Transactions on Signal Processidoi =
  {10.1109/IJCNN.2006.246722}ng}, vol.~40, pp.~967--969, April 1992.

\bibitem{Hayakawa2018}
D.~{Hayakawa}, T.~{Masuko}, and H.~{Fujimura}, ``Applying complex-valued neural
  networks to acoustic modeling for speech recognition,'' in {\em 2018
  Asia-Pacific Signal and Information Processing Association Annual Summit and
  Conference (APSIPA ASC)}, pp.~1725--1731, Nov 2018.

\bibitem{ACAR2018}
Y.~E. {ACAR}, M.~{CEYLAN}, and E.~{YALDIZ}, ``An examination on the effect of
  cvnn parameters while classifying the real-valued balanced and unbalanced
  data,'' in {\em 2018 International Conference on Artificial Intelligence and
  Data Processing (IDAP)}, pp.~1--5, Sep. 2018.

\bibitem{Ishizuka2018}
Y.~{Ishizuka}, S.~{Murai}, Y.~{Takahashi}, M.~{Kawai}, Y.~{Taniai}, and
  T.~{Naniwa}, ``Modeling walking behavior of powered exoskeleton based on
  complex-valued neural network,'' in {\em 2018 IEEE International Conference
  on Systems, Man, and Cybernetics (SMC)}, pp.~1927--1932, Oct 2018.

\bibitem{Yi2018a}
Q.~{Yi}, L.~{Xiao}, Y.~{Zhang}, B.~{Liao}, L.~{Ding}, and H.~{Peng},
  ``Nonlinearly activated complex-valued gradient neural network for complex
  matrix inversion,'' in {\em 2018 Ninth International Conference on
  Intelligent Control and Information Processing (ICICIP)}, pp.~44--48, Nov
  2018.

\bibitem{Cevik2018}
H.~H. {Çevik}, Y.~E. {Acar}, and M.~{Çunkaş}, ``Day ahead wind power
  forecasting using complex valued neural network,'' in {\em 2018 International
  Conference on Smart Energy Systems and Technologies (SEST)}, pp.~1--6, Sep.
  2018.

\bibitem{Gleich2018}
D.~{Gleich} and D.~{Sipos}, ``Complex valued convolutional neural network for
  terrasar-x patch categorization,'' in {\em EUSAR 2018; 12th European
  Conference on Synthetic Aperture Radar}, pp.~1--4, June 2018.

\bibitem{Gong2017}
W.~{Gong}, J.~{Liang}, and D.~{Li}, ``Design of high-capacity auto-associative
  memories based on the analysis of complex-valued neural networks,'' in {\em
  2017 International Workshop on Complex Systems and Networks (IWCSN)},
  pp.~161--168, Dec 2017.

\bibitem{Zhang2017b}
J.~{Zhang} and Y.~{Wu}, ``A new method for automatic sleep stage
  classification,'' {\em IEEE Transactions on Biomedical Circuits and Systems},
  vol.~11, pp.~1097--1110, Oct 2017.

\bibitem{Popa2017a}
C.~{Popa}, ``Complex-valued convolutional neural networks for real-valued image
  classification,'' in {\em 2017 International Joint Conference on Neural
  Networks (IJCNN)}, pp.~816--822, May 2017.

\bibitem{Liu2017}
S.~{Liu}, M.~{Xu}, J.~{Wang}, F.~{Lu}, W.~{Zhang}, H.~{Tian}, and G.~{Chang},
  ``A multilevel artificial neural network nonlinear equalizer for
  millimeter-wave mobile fronthaul systems,'' {\em Journal of Lightwave
  Technology}, vol.~35, pp.~4406--4417, Oct 2017.

\bibitem{Peker2016}
M.~{Peker}, B.~{Sen}, and D.~{Delen}, ``A novel method for automated diagnosis
  of epilepsy using complex-valued classifiers,'' {\em IEEE Journal of
  Biomedical and Health Informatics}, vol.~20, pp.~108--118, Jan 2016.

\bibitem{Amilia2015}
S.~{Amilia}, M.~D. {Sulistiyo}, and R.~N. {Dayawati}, ``Face image-based gender
  recognition using complex-valued neural network,'' in {\em 2015 3rd
  International Conference on Information and Communication Technology
  (ICoICT)}, pp.~201--206, May 2015.

\bibitem{Maeda2014}
Y.~{Maeda}, T.~{Fujiwara}, and H.~{Ito}, ``Robot control using high dimensional
  neural networks,'' in {\em 2014 Proceedings of the SICE Annual Conference
  (SICE)}, pp.~738--743, Sep. 2014.

\bibitem{Liu2014}
Y.~{Liu}, H.~{Huang}, and T.~{Huang}, ``Gain parameters based complex-valued
  backpropagation algorithm for learning and recognizing hand gestures,'' in
  {\em 2014 International Joint Conference on Neural Networks (IJCNN)},
  pp.~2162--2166, July 2014.

\bibitem{Olanrewaju2011}
R.~F. {Olanrewaju}, O.~{Khalifa}, A.~{Abdulla}, and A.~M.~Z. {Khedher},
  ``Detection of alterations in watermarked medical images using fast fourier
  transform and complex-valued neural network,'' in {\em 2011 4th International
  Conference on Mechatronics (ICOM)}, pp.~1--6, May 2011.

\bibitem{Haensch2010}
R.~{Haensch} and O.~{Hellwich}, ``Complex-valued convolutional neural networks
  for object detection in polsar data,'' in {\em 8th European Conference on
  Synthetic Aperture Radar}, pp.~1--4, June 2010.

\bibitem{Nait2010}
H.~{Nait-Charif}, ``Complex-valued neural networks fault tolerance in pattern
  classification applications,'' in {\em 2010 Second WRI Global Congress on
  Intelligent Systems}, vol.~3, pp.~154--157, Dec 2010.

\bibitem{Kitajima2010}
T.~{Kitajima} and T.~{Yasuno}, ``Output prediction of wind power generation
  system using complex-valued neural network,'' in {\em Proceedings of SICE
  Annual Conference 2010}, pp.~3610--3613, Aug 2010.

\bibitem{Ogawa2008}
S.~Fukami, T.~{Ogawa}, and H.~{Kanada}, ``Regularization for complex-valued
  network inversion,'' in {\em 2008 SICE Annual Conference}, pp.~1237--1242,
  Aug 2008.

\bibitem{Amin2008}
M.~F. {Amin}, M.~M. {Islam}, and K.~{Murase}, ``Single-layered complex-valued
  neural networks and their ensembles for real-valued classification
  problems,'' in {\em 2008 IEEE International Joint Conference on Neural
  Networks (IEEE World Congress on Computational Intelligence)},
  pp.~2500--2506, June 2008.

\bibitem{Nishikawa2005}
I.~{Nishikawa}, K.~{Sakakibara}, T.~{Iritani}, and Y.~{Kuroe}, ``2 types of
  complex-valued hopfield networks and the application to a traffic signal
  control,'' in {\em Proceedings. 2005 IEEE International Joint Conference on
  Neural Networks, 2005.}, vol.~2, pp.~782--787 vol. 2, July 2005.

\bibitem{Yadav2005}
A.~{Yadav}, D.~{Mishra}, S.~{Ray}, R.~N. {Yadav}, and P.~K. {Kalra},
  ``Representation of complex-valued neural networks: a real-valued approach,''
  in {\em Proceedings of 2005 International Conference on Intelligent Sensing
  and Information Processing, 2005.}, pp.~331--335, Jan 2005.

\bibitem{Kataoka1998}
M.~{Kataoka}, M.~{Kinouchi}, and M.~{Hagiwara}, ``Music information retrieval
  system using complex-valued recurrent neural networks,'' in {\em SMC'98
  Conference Proceedings. 1998 IEEE International Conference on Systems, Man,
  and Cybernetics (Cat. No.98CH36218)}, vol.~5, pp.~4290--4295 vol.5, Oct 1998.

\bibitem{Kinouchi1996}
M.~{Kinouchi} and M.~{Hagiwara}, ``Memorization of melodies by complex-valued
  recurrent network,'' in {\em Proceedings of International Conference on
  Neural Networks (ICNN'96)}, vol.~2, pp.~1324--1328 vol.2, June 1996.

\bibitem{Tan2020}
X.~{Tan}, M.~{Li}, P.~{Zhang}, Y.~{Wu}, and W.~{Song}, ``Complex-valued 3-d
  convolutional neural network for polsar image classification,'' {\em IEEE
  Geoscience and Remote Sensing Letters}, vol.~17, no.~6, pp.~1022--1026, 2020.

\bibitem{Georgiou1992}
G.~M. {Georgiou} and C.~{Koutsougeras}, ``Complex domain backpropagation,''
  {\em IEEE Transactions on Circuits and Systems II: Analog and Digital Signal
  Processing}, vol.~39, pp.~330--334, May 1992.

\bibitem{Hu2018a}
S.~{Hu}, S.~{Nagae}, and A.~{Hirose}, ``Millimeter-wave adaptive glucose
  concentration estimation with complex-valued neural networks,'' {\em IEEE
  Transactions on Biomedical Engineering}, pp.~1--1, 2018.

\bibitem{Hu2018b}
S.~{Hu} and A.~{Hirose}, ``Proposal of millimeter-wave adaptive
  glucose-concentration estimation system using complex-valued neural
  networks,'' in {\em IGARSS 2018 - 2018 IEEE International Geoscience and
  Remote Sensing Symposium}, pp.~4074--4077, July 2018.

\bibitem{Hata2016}
R.~{Hata} and K.~{Murase}, ``Multi-valued autoencoders for multi-valued neural
  networks,'' in {\em 2016 International Joint Conference on Neural Networks
  (IJCNN)}, pp.~4412--4417, July 2016.

\bibitem{Ding2014}
T.~{Ding} and A.~{Hirose}, ``Fading channel prediction based on combination of
  complex-valued neural networks and chirp z-transform,'' {\em IEEE
  Transactions on Neural Networks and Learning Systems}, vol.~25,
  pp.~1686--1695, Sep. 2014.

\bibitem{Suzuki2013}
Y.~{Suzuki} and M.~{Kobayashi}, ``Complex-valued bidirectional auto-associative
  memory,'' in {\em The 2013 International Joint Conference on Neural Networks
  (IJCNN)}, pp.~1--7, Aug 2013.

\bibitem{Mizote2013}
K.~{Mizote}, H.~{Kishikawa}, N.~{Goto}, and S.~{Yanagiya}, ``Optical label
  routing processing for bpsk labels using complex-valued neural network,''
  {\em Journal of Lightwave Technology}, vol.~31, pp.~1867--1876, June 2013.

\bibitem{Al_Nuaimi2012}
A.~Y.~H. {Al-Nuaimi}, M.~{Faijul Amin}, and K.~{Murase}, ``Enhancing mp3
  encoding by utilizing a predictive complex-valued neural network,'' in {\em
  The 2012 International Joint Conference on Neural Networks (IJCNN)},
  pp.~1--6, June 2012.

\bibitem{Hu2008}
J.~{Hu}, Z.~{Li}, Z.~{Hu}, D.~{Yao}, and J.~{Yu}, ``Spam detection with
  complex-valued neural network using behavior-based characteristics,'' in {\em
  2008 Second International Conference on Genetic and Evolutionary Computing},
  pp.~166--169, Sep. 2008.

\bibitem{Nishikawa2006}
I.~{Nishikawa}, T.~{Iritani}, and K.~{Sakakibara}, ``Improvements of the
  traffic signal control by complex-valued hopfield networks,'' in {\em The
  2006 IEEE International Joint Conference on Neural Network Proceedings},
  pp.~459--464, July 2006.

\bibitem{Nitta2018a}
T.~{Nitta} and Y.~{Kuroe}, ``Hyperbolic gradient operator and hyperbolic
  back-propagation learning algorithms,'' {\em IEEE Transactions on Neural
  Networks and Learning Systems}, vol.~29, pp.~1689--1702, May 2018.

\bibitem{Kominami2017}
Y.~{Kominami}, H.~{Ogawa}, and K.~{Murase}, ``Convolutional neural networks
  with multi-valued neurons,'' in {\em 2017 International Joint Conference on
  Neural Networks (IJCNN)}, pp.~2673--2678, May 2017.

\bibitem{Mandic2009}
D.~P. {Mandic}, ``Complex valued recurrent neural networks for noncircular
  complex signals,'' in {\em 2009 International Joint Conference on Neural
  Networks}, pp.~1987--1992, June 2009.

\bibitem{Hanna2002}
A.~I. {Hanna} and D.~P. {Mandic}, ``A normalised complex backpropagation
  algorithm,'' in {\em 2002 IEEE International Conference on Acoustics, Speech,
  and Signal Processing}, vol.~1, pp.~I--977--I--980, May 2002.

\bibitem{Adali2000}
T.~{Kim} and T.~{Adali}, ``Fully complex backpropagation for constant envelope
  signal processing,'' in {\em Neural Networks for Signal Processing X.
  Proceedings of the 2000 IEEE Signal Processing Society Workshop (Cat.
  No.00TH8501)}, vol.~1, pp.~231--240 vol.1, Dec 2000.

\bibitem{Kim2002}
T.~Kim and T.~Adali, ``Fully complex multi-layer perceptron network for
  nonlinear signal processing,'' {\em Journal of VLSI signal processing systems
  for signal, image and video technology}, vol.~32, pp.~29--43, Aug 2002.

\bibitem{Scardapane2018}
S.~{Scardapane}, S.~{Van Vaerenbergh}, A.~{Hussain}, and A.~{Uncini},
  ``Complex-valued neural networks with nonparametric activation functions,''
  {\em IEEE Transactions on Emerging Topics in Computational Intelligence},
  pp.~1--11, 2018.

\bibitem{Kobayashi2019}
M.~{Kobayashi}, ``Noise robust projection rule for hyperbolic hopfield neural
  networks,'' {\em IEEE Transactions on Neural Networks and Learning Systems},
  pp.~1--5, 2019.

\bibitem{Popa2018a}
C.~{Popa}, ``Complex-valued deep boltzmann machines,'' in {\em 2018
  International Joint Conference on Neural Networks (IJCNN)}, pp.~1--8, July
  2018.

\bibitem{Kobayashi2018c}
M.~{Kobayashi}, ``Stability of rotor hopfield neural networks with synchronous
  mode,'' {\em IEEE Transactions on Neural Networks and Learning Systems},
  vol.~29, pp.~744--748, March 2018.

\bibitem{Kuroe2002}
Y.~{Kuroe}, N.~{Hashimoto}, and T.~{Mori}, ``On energy function for
  complex-valued neural networks and its applications,'' in {\em Proceedings of
  the 9th International Conference on Neural Information Processing, 2002.
  ICONIP '02.}, vol.~3, pp.~1079--1083 vol.3, Nov 2002.

\bibitem{Yuan2019}
Q.~{Yuan}, D.~{Li}, Z.~{Wang}, C.~{Liu}, and C.~{He}, ``Channel estimation and
  pilot design for uplink sparse code multiple access system based on
  complex-valued sparse autoencoder,'' {\em IEEE Access}, pp.~1--1, 2019.

\bibitem{Li2019}
L.~{Li}, L.~G. {Wang}, F.~L. {Teixeira}, C.~{Liu}, A.~{Nehorai}, and T.~J.
  {Cui}, ``Deepnis: Deep neural network for nonlinear electromagnetic inverse
  scattering,'' {\em IEEE Transactions on Antennas and Propagation}, vol.~67,
  pp.~1819--1825, March 2019.

\bibitem{Gao2019}
J.~{Gao}, B.~{Deng}, Y.~{Qin}, H.~{Wang}, and X.~{Li}, ``Enhanced radar imaging
  using a complex-valued convolutional neural network,'' {\em IEEE Geoscience
  and Remote Sensing Letters}, vol.~16, pp.~35--39, Jan 2019.

\bibitem{Gu2018}
S.~{Gu} and L.~{Ding}, ``A complex-valued vgg network based deep learing
  algorithm for image recognition,'' in {\em 2018 Ninth International
  Conference on Intelligent Control and Information Processing (ICICIP)},
  pp.~340--343, Nov 2018.

\bibitem{Matlacz2018}
M.~{Matlacz} and G.~{Sarwas}, ``Crowd counting using complex convolutional
  neural network,'' in {\em 2018 Signal Processing: Algorithms, Architectures,
  Arrangements, and Applications (SPA)}, pp.~88--92, Sep. 2018.

\bibitem{Popa2018b}
C.~{Popa}, ``Deep hybrid real-complex-valued convolutional neural networks for
  image classification,'' in {\em 2018 International Joint Conference on Neural
  Networks (IJCNN)}, pp.~1--6, July 2018.

\bibitem{Shafran2018}
I.~{Shafran}, T.~{Bagby}, and R.~J. {Skerry-Ryan}, ``Complex evolution
  recurrent neural networks (cernns),'' in {\em 2018 IEEE International
  Conference on Acoustics, Speech and Signal Processing (ICASSP)},
  pp.~5854--5858, April 2018.

\bibitem{Lee2017}
Y.~{Lee}, C.~{Wang}, S.~{Wang}, J.~{Wang}, and C.~{Wu}, ``Fully complex deep
  neural network for phase-incorporating monaural source separation,'' in {\em
  2017 IEEE International Conference on Acoustics, Speech and Signal Processing
  (ICASSP)}, pp.~281--285, March 2017.

\bibitem{Lee2017b}
Y.-S. Lee, K.~Yu, S.-H. Chen, and J.-C. Wang, ``Discriminative training of
  complex-valued deep recurrent neural network for singing voice separation,''
  in {\em Proceedings of the 25th ACM International Conference on Multimedia},
  MM '17, (New York, NY, USA), pp.~1327--1335, ACM, 2017.

\bibitem{Kobayashi2019b}
M.~{Kobayashi}, ``O(2)-valued hopfield neural networks,'' {\em IEEE
  Transactions on Neural Networks and Learning Systems}, pp.~1--6, 2019.

\bibitem{Aizenberg2018a}
I.~{Aizenberg} and A.~{Gonzalez}, ``Image recognition using mlmvn and frequency
  domain features,'' in {\em 2018 International Joint Conference on Neural
  Networks (IJCNN)}, pp.~1--8, July 2018.

\bibitem{Aizenberg2018b}
I.~{Aizenberg} and Z.~{Khaliq}, ``Analysis of eeg using multilayer neural
  network with multi-valued neurons,'' in {\em 2018 IEEE Second International
  Conference on Data Stream Mining Processing (DSMP)}, pp.~392--396, Aug 2018.

\bibitem{Kobayashi2018b}
M.~{Kobayashi}, ``Decomposition of rotor hopfield neural networks using complex
  numbers,'' {\em IEEE Transactions on Neural Networks and Learning Systems},
  vol.~29, pp.~1366--1370, April 2018.

\bibitem{Virtue2017}
P.~{Virtue}, S.~X. {Yu}, and M.~{Lustig}, ``Better than real: Complex-valued
  neural nets for mri fingerprinting,'' in {\em 2017 IEEE International
  Conference on Image Processing (ICIP)}, pp.~3953--3957, Sep. 2017.

\bibitem{Ronghua2017}
J.~{Ronghua}, Z.~{Shulei}, Z.~{Lihua}, L.~{Qiuxia}, and I.~A. {Saeed},
  ``Prediction of soil moisture with complex-valued neural network,'' in {\em
  2017 29th Chinese Control And Decision Conference (CCDC)}, pp.~1231--1236,
  May 2017.

\bibitem{Aizenberg2017a}
I.~{Aizenberg}, A.~{Ordukhanov}, and F.~{O'Boy}, ``Mlmvn as an intelligent
  image filter,'' in {\em 2017 International Joint Conference on Neural
  Networks (IJCNN)}, pp.~3106--3113, May 2017.

\bibitem{Kobayashi2017}
M.~{Kobayashi}, ``Symmetric complex-valued hopfield neural networks,'' {\em
  IEEE Transactions on Neural Networks and Learning Systems}, vol.~28,
  pp.~1011--1015, April 2017.

\bibitem{Aizenberg2016a}
I.~{Aizenberg}, A.~{Luchetta}, S.~{Manetti}, and M.~C. {Piccirilli}, ``System
  identification using fra and a modified mlmvn with arbitrary complex-valued
  inputs,'' in {\em 2016 International Joint Conference on Neural Networks
  (IJCNN)}, pp.~4404--4411, July 2016.

\bibitem{Hacker2016}
C.~{Hacker}, I.~{Aizenberg}, and J.~{Wilson}, ``Gpu simulator of multilayer
  neural network based on multi-valued neurons,'' in {\em 2016 International
  Joint Conference on Neural Networks (IJCNN)}, pp.~4125--4132, July 2016.

\bibitem{Catelani2016}
M.~{Catelani}, L.~{Ciani}, A.~{Luchetta}, S.~{Manetti}, M.~C. {Piccirilli},
  A.~{Reatti}, and M.~K. {Kazimierczuk}, ``Mlmvnn for parameter fault detection
  in pwm dc-dc converters and its applications for buck dc-dc converter,'' in
  {\em 2016 IEEE 16th International Conference on Environment and Electrical
  Engineering (EEEIC)}, pp.~1--6, June 2016.

\bibitem{Aizenberg2014a}
E.~{Aizenberg} and I.~{Aizenberg}, ``Batch linear least squares-based learning
  algorithm for mlmvn with soft margins,'' in {\em 2014 IEEE Symposium on
  Computational Intelligence and Data Mining (CIDM)}, pp.~48--55, Dec 2014.

\bibitem{Pavaloiu2014}
I.~B. {Păvăloiu}, G.~{Dragoi}, and A.~{Vasile}, ``Gradient-descent training
  for phase-based neurons,'' in {\em 2014 18th International Conference on
  System Theory, Control and Computing (ICSTCC)}, pp.~874--878, Oct 2014.

\bibitem{Valle2014}
M.~E. {Valle}, ``An introduction to complex-valued recurrent correlation neural
  networks,'' in {\em 2014 International Joint Conference on Neural Networks
  (IJCNN)}, pp.~3387--3394, July 2014.

\bibitem{Aizenberg2013a}
I.~{Aizenberg}, ``A modified error-correction learning rule for multilayer
  neural network with multi-valued neurons,'' in {\em The 2013 International
  Joint Conference on Neural Networks (IJCNN)}, pp.~1--8, Aug 2013.

\bibitem{Wu2013}
S.~{Wu} and S.~{Lee}, ``Multi-valued neuron with new learning schemes,'' in
  {\em The 2013 International Joint Conference on Neural Networks (IJCNN)},
  pp.~1--7, Aug 2013.

\bibitem{Chen2013}
J.~{Chen}, S.~{Wu}, and S.~{Lee}, ``Modified learning for discrete multi-valued
  neuron,'' in {\em The 2013 International Joint Conference on Neural Networks
  (IJCNN)}, pp.~1--6, Aug 2013.

\bibitem{Aizenberg2011a}
I.~{Aizenberg}, S.~{Alexander}, and J.~{Jackson}, ``Recognition of blurred
  images using multilayer neural network based on multi-valued neurons,'' in
  {\em 2011 41st IEEE International Symposium on Multiple-Valued Logic},
  pp.~282--287, May 2011.

\bibitem{Aizenberg2008b}
I.~{Aizenberg}, D.~V. {Paliy}, J.~M. {Zurada}, and J.~T. {Astola}, ``Blur
  identification by multilayer neural network based on multivalued neurons,''
  {\em IEEE Transactions on Neural Networks}, vol.~19, pp.~883--898, May 2008.

\bibitem{Donq2001}
D.~L. Lee, ``Improving the capacity of complex-valued neural networks with a
  modified gradient descent learning rule,'' {\em IEEE Transactions on Neural
  Networks}, vol.~12, pp.~439--443, March 2001.

\bibitem{Aizenberg2000}
I.~{Aizenberg}, N.~{Aizenberg}, C.~{Butakov}, and E.~{Farberov}, ``Image
  recognition on the neural network based on multi-valued neurons,'' in {\em
  Proceedings 15th International Conference on Pattern Recognition. ICPR-2000},
  vol.~2, pp.~989--992 vol.2, Sep. 2000.

\bibitem{Aizenberg2000b}
I.~{Aizenberg}, N.~{Aizenberg}, T.~{Bregin}, C.~{Butakov}, and E.~{Farberov},
  ``Image processing using cellular neural networks based on multi-valued and
  universal binary neurons,'' in {\em Neural Networks for Signal Processing X.
  Proceedings of the 2000 IEEE Signal Processing Society Workshop (Cat.
  No.00TH8501)}, vol.~2, pp.~557--566 vol.2, Dec 2000.

\bibitem{Aizenberg1996}
N.~N. {Aizenberg}, I.~N. {Aizenberg}, and G.~A. {Krivosheev}, ``Multi-valued
  and universal binary neurons: mathematical model, learning, networks,
  application to image processing and pattern recognition,'' in {\em
  Proceedings of 13th International Conference on Pattern Recognition}, vol.~4,
  pp.~185--189 vol.4, Aug 1996.

\bibitem{Olga2014}
O.~Fink, E.~Zio, and U.~Weidmann, ``Predicting component reliability and level
  of degradation with complex-valued neural networks,'' {\em Reliability
  Engineering \& System Safety}, vol.~121, pp.~198 -- 206, 2014.

\bibitem{Jankowski1996}
S.~{Jankowski}, A.~{Lozowski}, and J.~M. {Zurada}, ``Complex-valued multistate
  neural associative memory,'' {\em IEEE Transactions on Neural Networks},
  vol.~7, pp.~1491--1496, Nov 1996.

\bibitem{Tanaka2009}
G.~Tanaka and K.~Aihara, ``Complex-valued multistate associative memory with
  nonlinear multilevel functions for gray-level image reconstruction,'' {\em
  Trans. Neur. Netw.}, vol.~20, pp.~1463--1473, Sept. 2009.

\bibitem{Aizenberg2006}
I.~Aizenberg, P.~Ruusuvuori, O.~Yli-Harja, and J.~Astola, ``Multilayer neural
  network based on multi-valued neurons (mlmvn) applied to classification of
  microarray gene expression data,'' pp.~27--30, 07 2006.

\bibitem{Ding2019}
L.~{Ding}, L.~{Xiao}, K.~{Zhou}, Y.~{Lan}, Y.~{Zhang}, and J.~{Li}, ``An
  improved complex-valued recurrent neural network model for time-varying
  complex-valued sylvester equation,'' {\em IEEE Access}, vol.~7,
  pp.~19291--19302, 2019.

\bibitem{Marseet2017}
A.~{Marseet} and F.~{Sahin}, ``Application of complex-valued convolutional
  neural network for next generation wireless networks,'' in {\em 2017 IEEE
  Western New York Image and Signal Processing Workshop (WNYISPW)}, pp.~1--5,
  Nov 2017.

\bibitem{Wilmanski2016}
M.~{Wilmanski}, C.~{Kreucher}, and A.~{Hero}, ``Complex input convolutional
  neural networks for wide angle sar atr,'' in {\em 2016 IEEE Global Conference
  on Signal and Information Processing (GlobalSIP)}, pp.~1037--1041, Dec 2016.

\bibitem{Aizenberg1973}
N.~N. Aizenberg, Y.~L. Ivas'kiv, D.~A. Pospelov, and G.~F. Khudyakov,
  ``Multivalued threshold functions,'' {\em Cybernetics}, vol.~9, pp.~61--77,
  Jan 1973.

\bibitem{Ajzenberg1972}
N.~N. Ajzenberg and Živko Tošić, ``A generalization of the threshold
  functions,'' {\em Publikacije Elektrotehničkog fakulteta. Serija Matematika
  i fizika}, no.~381/409, pp.~97--99, 1972.

\bibitem{Clarke1990}
T.~L. {Clarke}, ``Generalization of neural networks to the complex plane,'' in
  {\em 1990 IJCNN International Joint Conference on Neural Networks},
  pp.~435--440 vol.2, June 1990.

\bibitem{Kuroe2003}
Y.~Kuroe, M.~Yoshid, and T.~Mori, ``On activation functions for complex-valued
  neural networks --- existence of energy functions ---,'' in {\em Artificial
  Neural Networks and Neural Information Processing --- ICANN/ICONIP 2003}
  (O.~Kaynak, E.~Alpaydin, E.~Oja, and L.~Xu, eds.), (Berlin, Heidelberg),
  pp.~985--992, Springer Berlin Heidelberg, 2003.

\bibitem{Noest1988}
A.~J. Noest, ``Associative memory in sparse phasor neural networks,'' {\em
  Europhysics Letters ({EPL})}, vol.~6, pp.~469--474, jul 1988.

\bibitem{Miyajima2000}
T.~Miyajima, F.~Baisho, and K.~Yamanaka, ``A phasor model with resting
  states,'' {\em IEICE Transactions on Information and Systems}, vol.~E83D,
  pp.~299--301, 02 2000.

\bibitem{Hirose1992b}
A.~{Hirose}, ``Dynamics of fully complex-valued neural networks,'' {\em
  Electronics Letters}, vol.~28, pp.~1492--1494, July 1992.

\bibitem{Nemoto2002}
I.~Nemoto and K.~Saito, ``A complex-valued version of nagumo--sato model of a
  single neuron and its behavior,'' {\em Neural Netw.}, vol.~15, pp.~833--853,
  Sept. 2002.

\bibitem{Zemel1995}
R.~S. Zemel, C.~K. Williams, and M.~C. Mozer, ``Lending direction to neural
  networks,'' {\em Neural Networks}, vol.~8, no.~4, pp.~503 -- 512, 1995.

\bibitem{Aizenberg1995}
N.~N. Aizenberg, I.~N. Aizenberg, and G.~A. Krivosheev, ``Multi-valued neurons:
  Learning, networks, application to image recognition and extrapolation of
  temporal series,'' in {\em From Natural to Artificial Neural Computation}
  (J.~Mira and F.~Sandoval, eds.), (Berlin, Heidelberg), pp.~389--395, Springer
  Berlin Heidelberg, 1995.

\bibitem{Kim2001}
T.~Kim and T.~Adali, ``Complex backpropagation neural network using elementary
  transcendental activation functions,'' vol.~2, pp.~1281 -- 1284 vol.2, 02
  2001.

\bibitem{Kim2003}
T.~Kim and T.~Adali, ``Approximation by fully complex multilayer perceptrons,''
  {\em Neural Comput.}, vol.~15, p.~1641–1666, July 2003.

\bibitem{Inhyok1995}
I.~Cha and S.~A. {Kassam}, ``Channel equalization using adaptive complex radial
  basis function networks,'' {\em IEEE Journal on Selected Areas in
  Communications}, vol.~13, pp.~122--131, Jan 1995.

\bibitem{Chen1994a}
S.~Chen, S.~McLaughlin, and B.~Mulgrew, ``Complex-valued radial basis function
  network, part i: Network architecture and learning algorithms,'' {\em Signal
  Process.}, vol.~35, p.~19–31, Jan. 1994.

\bibitem{Chen1994b}
S.~Chen, S.~McLaughlin, and B.~Mulgrew, ``Complex-valued radial basis function
  network, part ii: Application to digital communications channel
  equalisation,'' {\em Signal Process.}, vol.~36, p.~175–188, Mar. 1994.

\bibitem{Deng2002}
D.~Jianping, N.~Sundararajan, and P.~{Saratchandran}, ``Communication channel
  equalization using complex-valued minimal radial basis function neural
  networks,'' {\em IEEE Transactions on Neural Networks}, vol.~13,
  pp.~687--696, May 2002.

\bibitem{Uncini1999}
A.~{Uncini}, L.~{Vecci}, P.~{Campolucci}, and F.~{Piazza}, ``Complex-valued
  neural networks with adaptive spline activation function for
  digital-radio-links nonlinear equalization,'' {\em IEEE Transactions on
  Signal Processing}, vol.~47, pp.~505--514, Feb 1999.

\bibitem{Scarpiniti2008}
M.~Scarpiniti, D.~Vigliano, R.~Parisi, and A.~Uncini, ``Generalized splitting
  functions for blind separation of complex signals,'' {\em Neurocomput.},
  vol.~71, pp.~2245--2270, June 2008.

\bibitem{Hahnloser2000}
R.~Hahnloser, R.~Sarpeshkar, M.~Mahowald, R.~Douglas, and H.~Seung, ``Digital
  selection and analogue amplification coexist in a cortex-inspired silicon
  circuit,'' {\em Nature}, vol.~405, pp.~947--51, 07 2000.

\bibitem{Trabelsi2017}
C.~Trabelsi, O.~Bilaniuk, Y.~Zhang, D.~Serdyuk, S.~Subramanian, J.~F. Santos,
  S.~Mehri, N.~Rostamzadeh, Y.~Bengio, and C.~J. Pal, ``Deep complex
  networks,'' {\em CoRR}, vol.~abs/1705.09792, 2018.

\bibitem{Savitha2013}
R.~{Savitha}, S.~{Suresh}, and N.~{Sundararajan}, ``Projection-based fast
  learning fully complex-valued relaxation neural network,'' {\em IEEE
  Transactions on Neural Networks and Learning Systems}, vol.~24, pp.~529--541,
  April 2013.

\bibitem{Kim1990}
M.~S. {Kim} and C.~C. {Guest}, ``Modification of backpropagation networks for
  complex-valued signal processing in frequency domain,'' in {\em 1990 IJCNN
  International Joint Conference on Neural Networks}, pp.~27--31 vol.3, June
  1990.

\bibitem{Chen2000}
R.-C. Huang and M.-S. Chen, ``Adaptive equalization using complex-valued
  multilayered neural network based on the extended kalman filter,'' vol.~1,
  pp.~519 -- 524 vol.1, 02 2000.

\bibitem{Ceylan2005}
M.~Ceylan, N.~Çetinkaya, R.~Ceylan, and Y.~Özbay, ``Comparison of
  complex-valued neural network and fuzzy clustering complex-valued neural
  network for load-flow analysis,'' vol.~3949, pp.~92--99, 01 2005.

\bibitem{Tay2007}
C.~Tay, K.~Tanizawa, and A.~Hirose, ``Error reduction in holographic movies
  using a hybrid learning method in coherent neural networks,'' vol.~47,
  pp.~884--893, 09 2007.

\bibitem{Tsuzuki2013}
H.~Tsuzuki, M.~Kugler, S.~Kuroyanagi, and A.~Iwata, ``An approach for sound
  source localization by complex-valued neural network,'' {\em IEICE
  Transactions on Information and Systems}, vol.~E96.D, pp.~2257--2265, 10
  2013.

\bibitem{Zhang2016}
H.~{Zhang} and D.~P. {Mandic}, ``Is a complex-valued stepsize advantageous in
  complex-valued gradient learning algorithms?,'' {\em IEEE Transactions on
  Neural Networks and Learning Systems}, vol.~27, pp.~2730--2735, Dec 2016.

\bibitem{Popovic2004}
and D.~H.~{Popovic} and D.~P. {Mandic}, ``Complex-valued estimation of wind
  profile and wind power,'' in {\em Proceedings of the 12th IEEE Mediterranean
  Electrotechnical Conference (IEEE Cat. No.04CH37521)}, vol.~3, pp.~1037--1040
  Vol.3, May 2004.

\bibitem{Aizenberg2007}
I.~Aizenberg and C.~Moraga, ``Multilayer feedforward neural network based on
  multi-valued neurons (mlmvn) and a backpropagation learning algorithm,'' {\em
  Soft Comput.}, vol.~11, pp.~169--183, 01 2007.

\bibitem{Shamima2013}
B.~{Shamima}, R.~{Savitha}, S.~{Suresh}, and S.~{Saraswathi}, ``Protein
  secondary structure prediction using a fully complex-valued relaxation
  network,'' in {\em The 2013 International Joint Conference on Neural Networks
  (IJCNN)}, pp.~1--8, Aug 2013.

\bibitem{Aizenberg2008}
I.~Aizenberg, D.~V. Paliy, J.~M. Zurada, and J.~T. Astola, ``Blur
  identification by multilayer neural network based on multivalued neurons,''
  {\em Trans. Neur. Netw.}, vol.~19, pp.~883--898, May 2008.

\bibitem{Aizenberg2000c}
I.~Aizenberg, N.~Aizenberg, T.~Bregin, C.~Butakoff, E.~Farberov, N.~Merzlyakov,
  and O.~Milukova, ``Blur recognition on the neural network based on
  multi-valued neurons,'' {\em Journal of Image and Graphics. Proc. Intl. Conf.
  on Image and Graphics (ICIG)}, vol.~5, pp.~127--130, 01 2000.

\bibitem{Leung1991}
H.~{Leung} and S.~{Haykin}, ``The complex backpropagation algorithm,'' {\em
  IEEE Transactions on Signal Processing}, vol.~39, pp.~2101--2104, Sep. 1991.

\bibitem{van1994}
A.~{van den Bos}, ``Complex gradient and hessian,'' {\em IEE Proceedings -
  Vision, Image and Signal Processing}, vol.~141, pp.~380--383, Dec 1994.

\bibitem{Haykin2014}
S.~Haykin and S.~Haykin, {\em Adaptive Filter Theory}.
\newblock Pearson, 2014.

\bibitem{Goh2008}
S.~Lee~Goh and D.~P~Mandic, ``An augmented crtrl for complex-valued recurrent
  neural networks,'' {\em Neural networks : the official journal of the
  International Neural Network Society}, vol.~20, pp.~1061--6, 01 2008.

\bibitem{Hirose1999}
A.~Hirose and H.~Onishi, ``Proposal of relative minimization learning
  forbehavior stabilization of complex-valued recurrent neural networks,''
  vol.~24(1-3), pp.~163--171, Elsevier, 1999.

\bibitem{Mandic2008}
D.~Mandic, S.~Javidi, S.~Goh, A.~Kuh, and K.~Aihara, ``Complex-valued
  prediction of wind profile using augmented complex statistics,'' {\em
  Renewable Energy}, vol.~34, 08 2008.

\bibitem{Solazzi2002}
M.~Solazzi, A.~Uncini, E.~Di~Claudio, and R.~Parisi, ``Complex discriminative
  learning bayesian neural equalizer,'' {\em Signal Processing}, vol.~81,
  pp.~2493--2502, 10 2002.

\bibitem{Benvenuto1991}
N.~Benvenuto, M.~Marchesi, F.~Piazza, and A.~Uncini, ``Non linear satellite
  radio links equalized using blind neural networks,'' pp.~1521 -- 1524 vol.3,
  05 1991.

\bibitem{Suksmono2003}
A.~B. Suksmono and A.~Hirose, ``Adaptive beamforming by using complex-valued
  multi layer perceptron", booktitle="artificial neural networks and neural
  information processing --- icann/iconip 2003,'' (Berlin, Heidelberg),
  pp.~959--966, Springer Berlin Heidelberg, 2003.

\bibitem{Hirose2000}
A.~Hirose and M.~Kiuchi, ``Coherent optical associative memory system that
  processes complex-amplitude information,'' {\em Photonics Technology Letters,
  IEEE}, vol.~12, pp.~564 -- 566, 06 2000.

\bibitem{Oyama2018}
K.~{Oyama} and A.~{Hirose}, ``Adaptive phase-singular-unit restoration with
  entire-spectrum-processing complex-valued neural networks in interferometric
  sar,'' {\em Electronics Letters}, vol.~54, no.~1, pp.~43--45, 2018.

\bibitem{Chistyakov2011}
Y.~Chistyakov, E.~Kholodova, A.~Minin, H.~Zimmermann, and A.~Knoll, ``Modeling
  of electric power transformer using complex-valued neural networks,'' {\em
  Energy Procedia}, vol.~12, p.~638–647, 12 2011.

\bibitem{Chistyakov2012}
A.~Minin, Y.~Chistyakov, E.~Kholodova, H.~Zimmermann, and A.~Knoll,
  ``Complex-valued open recurrent neural network for power transformer
  modeling,'' {\em Int. J. Appl. Math. Inform}, vol.~6, pp.~41--48, 01 2012.

\bibitem{Miyauchi1993}
M.~Miyauchi, M.~Seki, A.~Watanabe, and A.~Miyauchi, ``Interpretation of optical
  flow through complex neural network,'' in {\em New Trends in Neural
  Computation} (J.~Mira, J.~Cabestany, and A.~Prieto, eds.), (Berlin,
  Heidelberg), pp.~645--650, Springer Berlin Heidelberg, 1993.

\bibitem{Miyauchi1992}
M.~{Miyauchi} and M.~{Seki}, ``Interpretation of optical flow through neural
  network learning,'' in {\em [Proceedings] Singapore ICCS/ISITA `92},
  pp.~1247--1251 vol.3, 1992.

\bibitem{Hirose2006b}
A.~Hirose, T.~Higo, and K.~Tanizawa, ``Holographic three-dimensional movie
  generation with frame interpolation using coherent neural networks,'' pp.~492
  -- 497, 01 2006.

\bibitem{Hirose2006a}
A.~Hirose, T.~Higo, and K.~Tanizawa, ``Efficient generation of holographic
  movies with frame interpolation using a coherent neural network,'' {\em Ieice
  Electronic Express}, vol.~3, pp.~417--423, 10 2006.

\bibitem{Yao2019}
X.~{Yao}, X.~{Shi}, and F.~{Zhou}, ``Complex-value convolutional neural network
  for classification of human activities,'' in {\em 2019 6th Asia-Pacific
  Conference on Synthetic Aperture Radar (APSAR)}, pp.~1--6, 2019.

\bibitem{Yao2020}
X.~{Yao}, X.~{Shi}, and F.~{Zhou}, ``Human activities classification based on
  complex-value convolutional neural network,'' {\em IEEE Sensors Journal},
  vol.~20, no.~13, pp.~7169--7180, 2020.

\bibitem{Dong2019}
T.~{Dong} and T.~{Huang}, ``Neural cryptography based on complex-valued neural
  network,'' {\em IEEE Transactions on Neural Networks and Learning Systems},
  pp.~1--6, 2019.

\bibitem{Rahman2001}
A.~F. Rahman, W.~G. Howells, and M.~C. Fairhurst, ``A multiexpert framework for
  character recognition: A novel application of clifford networks,'' {\em
  Trans. Neur. Netw.}, vol.~12, pp.~101--112, Jan. 2001.

\bibitem{Sun2019}
Q.~{Sun}, X.~{Li}, L.~{Li}, X.~{Liu}, F.~{Liu}, and L.~{Jiao},
  ``Semi-supervised complex-valued gan for polarimetric sar image
  classification,'' in {\em IGARSS 2019 - 2019 IEEE International Geoscience
  and Remote Sensing Symposium}, pp.~3245--3248, 2019.

\bibitem{Anderson2017}
T.~Anderson, ``Split complex convolutional neural networks,'' 2017.

\bibitem{Joshua2019}
J.~{Bassey}, X.~{Li}, and L.~{Qian}, ``An experimental study of multi-layer
  multi-valued neural network,'' in {\em 2019 2nd International Conference on
  Data Intelligence and Security (ICDIS)}, pp.~233--236, 2019.

\bibitem{Nils2018}
N.~Monning and S.~Manandhar, ``Evaluation of complex-valued neural networks on
  real-valued classification tasks,'' {\em CoRR}, vol.~abs/1811.12351, 2018.

\end{thebibliography}
\end{document}